\documentclass[letterpaper, 10 pt, conference]{ieeeconf}  % Comment this line out if you need a4paper

\IEEEoverridecommandlockouts                              % This command is only needed if 
\usepackage{graphics} % for pdf, bitmapped graphics files
\usepackage{epsfig} % for postscript graphics files
\usepackage{times} % assumes new font selection scheme installed
\usepackage{amsmath} % assumes amsmath package installed
\usepackage{amsfonts}
\usepackage{amssymb}
\renewcommand\QED{\QEDopen}

\usepackage{tikz}
\usetikzlibrary{matrix, positioning, calc}
\usepackage{standalone}
\usepackage{booktabs}
\usepackage{graphbox}

\usepackage{subcaption}
\usepackage{multirow,bigstrut}
\usepackage{tabu}
\usepackage{lipsum}
\usepackage{balance}

\usepackage{url}

\title{\LARGE \bf
Equivariant Ensembles and Regularization for Reinforcement Learning in Map-based Path Planning
}

\author{Mirco Theile$^{1}$, Hongpeng Cao$^{1}$, Marco Caccamo$^{1}$, and Alberto L. Sangiovanni-Vincentelli$^{2}$% <-this % stops a space
\thanks{Marco Caccamo was supported by an Alexander von Humboldt Professorship endowed by the German Federal Ministry of Education and Research.}% <-this % stops a space
\thanks{$^{1}$Mirco Theile, Hongpeng Cao, and Marco Caccamo are with TUM School of Engineering and Design, Technical University of Munich, Germany
        {\tt\small mirco.theile,cao.hongpeng,mcaccamo@tum.de}}%
\thanks{$^{2}$Alberto L. Sangiovanni-Vincentelli is with Dept. of Electrical Engineering and Computer Sciences, University of California, Berkeley, USA
        {\tt\small alberto@berkeley.edu}}%
}

\begin{document}

\maketitle
\thispagestyle{empty}
\pagestyle{empty}

%%%%%%%%%%%%%%%%%%%%%%%%%%%%%%%%%%%%%%%%%%%%%%%%%%%%%%%%%%%%%%%%%%%%%%%%%%%%%%%%
\begin{abstract}
In reinforcement learning (RL), exploiting environmental symmetries can significantly enhance efficiency, robustness, and performance. However, ensuring that the deep RL policy and value networks are respectively equivariant and invariant to exploit these symmetries is a substantial challenge. Related works try to design networks that are equivariant and invariant by construction, limiting them to a very restricted library of components, which in turn hampers the expressiveness of the networks. This paper proposes a method to construct equivariant policies and invariant value functions without specialized neural network components, which we term equivariant ensembles. We further add a regularization term for adding inductive bias during training. In a map-based path planning case study, we show how equivariant ensembles and regularization benefit sample efficiency and performance.
\end{abstract}

%%%%%%%%%%%%%%%%%%%%%%%%%%%%%%%%%%%%%%%%%%%%%%%%%%%%%%%%%%%%%%%%%%%%%%%%%%%%%%%%
\section{Introduction}

\begin{figure}
    \centering
    \resizebox{\columnwidth}{!}{\input{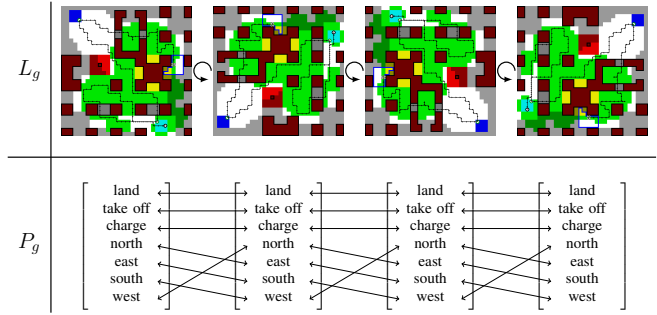}}
    \caption{Visualization of the equivariances in a UAV coverage path planning application showing the input and output transformations $L_g$ and $P_g$ for all rotations in $G$.}
    \vspace{-3pt}
    \label{fig:transforms}
\end{figure}

Reinforcement learning (RL) is a rapidly advancing methodology for learning policies through interactions with environments, as it promises to address complex real-world problems that were previously unsolvable. Its application breadth is continuously widening with applications in control, planning, and general optimization domains. Many of these environments contain symmetries that could be exploited for improved training efficiency, robustness, and performance. Symmetries in the environments result in equivariance (see Figure~\ref{fig:transforms} for an example) and invariance properties of the optimal policy and value function, respectively. Therefore, imposing these properties on the learning components should be beneficial.

Inspired by the success of convolutional neural networks that make neural networks equivariant to translations \cite{krizhevsky2012imagenet}, earlier research focused on designing neural networks such that they are equivariant or invariant to the symmetry transformations \cite{van2020mdp, wang2021mathrm}. However, this endeavor can be daunting as it limits the neural network components to only small subsets of equivariant and invariant layers, opposing the trend toward ever-increasing complexity of neural networks for performance improvements. Recently, related research has shifted towards adding a regularization term to the training loss that nudges the networks towards equivariance and invariance without constraining the design choices for the networks \cite{raileanu2021automatic, nguyen2024symmetry}.

In this paper, we incorporate symmetries into the learning process by constructing equivariant policies and invariant value functions without the need for special neural network designs. We introduce \textit{equivariant ensembles} that average over the networks' outputs for all symmetry transformations. We prove that policy and value ensembles are equivariant and invariant, respectively, and show how they enrich the gradients in policy optimization algorithms such as proximal policy optimization (PPO) \cite{schulman2017proximal}. We further use regularization to push the individual components toward the ensembles, adding inductive bias.

To showcase the benefits of the equivariant ensembles and regularization, we evaluate their performance in a challenging, long-horizon, map-based planning application, the unmanned aerial vehicle (UAV) Coverage Path Planning (CPP) problem \cite{theile2023learning}. In this case study, the environment state can be represented as a map (\cite{theile2020uav}), and rotational symmetries can be exploited, as visualized in Figure~\ref{fig:transforms}. The results show that the ensemble makes the policy equivariant and that combining the ensemble and regularization improves performance significantly. We further show that regularization on the policy does not guarantee equivariance, which should be considered when regularizing the value estimate towards invariance.

To summarize, the contributions of this paper are as follows:
\begin{itemize}
    \item Introduction of equivariant ensembles to enforce equivariance and invariance of policies and value functions without special neural network designs;
    \item Combination of equivariant ensembles and regularization to enrich the gradients in policy optimization algorithms through implicit data augmentation and providing inductive bias;
    \item Implementation\footnote{Code: \url{https://github.com/theilem/uavSim.git}} of ensembles and regularization in the long-horizon problem of UAV coverage path planning;
    \item Analyzing the effects of ensembles and regularization on sample efficiency, performance, and out-of-distribution generalization.
\end{itemize}
\section{Preliminaries}

\subsection{Invariance and Equivariance}

A function $f: \mathcal{X} \to \mathcal{Y}$ is defined to be \textit{invariant} to transformations from a group of linear transformations $G$ if for all $g\in G$ and their corresponding transformation operator $L_g: \mathcal{X}\to\mathcal{X}$ the following equality holds:
\begin{equation}
    f(x) = f(L_g[x]), \quad \forall g\in G, x\in\mathcal{X}.
\end{equation}
In this case, the function output does not change when applying the transformations on the function domain. The group $G$ is characterized by containing the identity, the inverse of each element, and the compositions of all pairs of elements, i.e., it is closed.

Adding the corresponding linear transformation operators $K_g: \mathcal{Y}\to\mathcal{Y}$ for all $g\in G$, the function $f$ is \textit{equivariant} to the transformations of $G$ if 
\begin{equation}
    K_g[f(x)] = f(L_g[x]), \quad \forall g\in G, x\in\mathcal{X}
\end{equation}

Consider a stochastic function $p:\mathcal{X}\to \mathcal{P}(\mathcal{Y})$ that maps $\mathcal{X}$ to the space of all probability distributions over $\mathcal{Y}$. We define the transformation operator $P_g: \mathcal{P}(\mathcal{Y})\to \mathcal{P}(\mathcal{Y})$ to which $p$ is equivariant if 
\begin{equation}\label{eq:equi_prob}
    P_g[p(\cdot|x)] = p(\cdot|L_g[x]), \quad \forall g\in G, x\in\mathcal{X}.
\end{equation}
Given the equivalence in \eqref{eq:equi_prob}, from equivariance of the stochastic function w.r.t. the transformation $P_g$ it follows that
\begin{equation}
    p(y | x) = p(K_g[y]~|~L_g[x]), \quad \forall g\in G, x\in\mathcal{X}, y\in\mathcal{Y},
\end{equation}
giving a relationship between $P_g$ and $K_g$.

\subsection{Reinforcement Learning}

\subsubsection{Fundamentals}
Reinforcement learning (RL) (\cite{sutton1998introduction}) aims to find a policy $\pi$ that maximizes the cumulative discounted reward of a Markov Decision Process (MDP). An MDP is defined through the tuple $(\mathcal{S}, \mathcal{A}, \mathrm{P}, \mathrm{R}, \gamma)$, with the state space $\mathcal{S}$, action space $\mathcal{A}$, probabilistic transition function $\mathrm{P}:\mathcal{S}\times\mathcal{A}\to \mathcal{P}(\mathcal{S})$, reward function $\mathrm{R}:\mathcal{S}\times\mathcal{A}\to\mathbb{R}$, and discount factor $\gamma\in [0, 1]$. The $\mathcal{P}(\mathcal{S})$ in the transition function stands for the space of all probability distributions over $\mathcal{S}$. The cumulative discounted reward, called return or value function, is defined as 
\begin{equation}\label{eq:value}
    \mathrm{V}^\pi(s) = \mathbb{E}\left[ \sum_{t=0}^\infty \gamma^t \mathrm{R}(s_t, a_t) ~\Big{|}~ s_{t+1} \sim \mathrm{P}, a_t \sim \pi, s_0 = s \right],
\end{equation}
when following a stochastic policy $\pi:\mathcal{S}\to\mathcal{P}(\mathcal{A})$. The state-action value function, the Q-value, is defined recursively as
\begin{equation}\label{eq:q_value}
    \mathrm{Q}^\pi(s,a) = \mathrm{R}(s,a) + \gamma \mathbb{E}_{s' \sim \mathrm{P}, a'\sim \pi}\left[ \mathrm{Q}^\pi(s',a')\right],
\end{equation}
in which $s'$ and $a'$ are the next state and action, respectively, which is the common notation adopted in the following. The Q-value determines the value of a specific action at the given state if following the policy afterward. Therefore, the advantage is defined as 
\begin{equation}
    \mathrm{A}^\pi(s,a) = \mathrm{Q}^\pi(s,a) - \mathrm{V}^\pi(s),
\end{equation}
indicating how much better or worse is a specific action at a given state compared to the expectation when following $\pi$.

\subsubsection{Proximal Policy Optimization (PPO)}
A popular deep RL algorithm is PPO \cite{schulman2017proximal}, which trains an actor network for the policy $\pi_\phi$ with parameters $\phi$ and a critic network to approximate the state-value function $\mathrm{V}^\pi_\theta$ with parameters $\theta$. The key underlying equation to train the actor is the policy optimization (PO) objective given by
\begin{equation}\label{eq:po_loss}
    J_\text{PO}(\phi) = \mathbb{E}_{(s, a)\sim\tau^{\pi_{\phi_\text{old}}}}\left[\frac{\pi_\phi(a|s)}{\pi_{\phi_\text{old}}(a|s)}\mathrm{A}^{\pi_{\phi_\text{old}}}(s,a)\right],
\end{equation}
in which $\pi_{\phi_\text{old}}$ is the \textit{behavior} policy, which in PPO is the previous $\pi_\phi$ that was used to collect trajectories called a rollout $\tau^{\pi_{\phi_\text{old}}}$. The advantage $\mathrm{A}^{\pi_{\phi_\text{old}}}$ is estimated using the generalized advantage estimate (GAE) \cite{schulman2015high}, which uses the discounted cumulative reward observed during the rollouts and the value estimate from the critic. In essence, the PO objective increases the probability of actions with positive advantage while decreasing it for actions with negative advantage. PPO adapts this objective by constraining the allowed difference between $\pi_\phi$ and $\pi_{\phi_\text{old}}$. The critic is trained on the discounted cumulative reward observed during the rollouts. Therefore, PPO is an on-policy RL algorithm since it can only be trained on data gathered by the behavior policy.

\subsubsection{Invariance and Equivariance in RL}
Symmetrical MDPs have the following equivalences
\begin{align}
    \mathrm{P}(L_g[s']|L_g[s], K_g[a]) =& \mathrm{P}(s'|s,a), \quad\forall g\in G \\
    \mathrm{R}(L_g[s], K_g[a]) =& \mathrm{R}(s,a), \quad\forall g\in G,
\end{align}
i.e., invariant transition and reward functions. It is shown that for symmetrical MDPs, the optimal policy $\pi^*$ is equivariant \cite{wang2021mathrm}, yielding
\begin{equation}
    P_g[\pi^*(\cdot | s)] = \pi^*(\cdot | L_g[s]), \quad\forall g\in G.
\end{equation}
Additionally, if a policy $\pi_\text{eq}$ is equivariant, the corresponding value function is invariant, i.e.,
\begin{equation}
    \mathrm{V}^{\pi_\text{eq}}(L_g[s]) = \mathrm{V}^{\pi_\text{eq}}(s), \quad\forall g\in G.
\end{equation}

\section{Related Research}
In this section, we review existing literature on the exploitation of symmetries in RL and highlight our contribution.

\subsection{Equivariant Neural Networks}
Equivariant neural networks directly embed symmetries in structures of the neural networks to constrain inputs and outputs to satisfy equivariance requirements. The critical components of equivariant networks are \textit{equivariant layers}, such as equivariant MLP \cite{van2020mdp} and special CNN layers \cite{cohen2019general} whose weights are designed to satisfy equivariance constraints. The idea of the equivariant network is applied to many state-based tasks such as classical control \cite{van2020mdp, finzi2021residual} and vision-based tasks such as robotic manipulation \cite{wang2022equivariant, wang2021mathrm, wang2022surprising}, showing significant improvement in sample efficiency. Nguyen et al.~\cite{nguyen2023equivariant} extend the application of equivariant neural networks from standard MDPs to Partially Observable Markov Decision Processes (POMDPs). However, designing equivariant models is non-trivial as it needs a deep understanding of neural networks and it constrains the choice of advanced network architectures. Furthermore, equivariant models may introduce instabilities in the training process compared to standard architectures \cite{finzi2021residual}.

\subsection{Regularization}

Regularization-based approaches incorporate the invariance or equivariance properties as auxiliary terms in the objective functions for the training of the policy and value functions~\cite{raileanu2021automatic, yarats2020image, nguyen2024symmetry}. Invariance regularization for the policy and value is typically conducted in data augmentation \cite{raileanu2021automatic, yarats2020image, laskin2020reinforcement} with the underlying idea that the value and action should be the same when applying observation transformations. The transformations can include photometric augmentation in vision-based observations or noise injections in state-based observations. The regularization of the policy minimizes the Kullback–Leibler (KL) divergence \cite{raileanu2021automatic} of the action distribution between the original state and augmented states.

In many cases, the policy should be equivariant under some state transformations, i.e., the policy should also be transformed. To regularize the policy to be equivariant, \cite{nguyen2024symmetry} proposes to minimize the mean-squared error loss between the means of policy on the original state and the transformed policies on transformed states in the SAC algorithm \cite{haarnoja2018soft} in a robotic manipulation task. 

\subsection{Our work}
We propose equivariant policy ensembles for policy and value function regularization to exploit symmetries in reinforcement learning. Compared to the equivariant neural networks \cite{van2020mdp, wang2022equivariant, wang2021mathrm, wang2022surprising,nguyen2023equivariant}, our approach does not need special neural network designs and can be easily integrated with existing standard reinforcement learning algorithms. Compared with the regularization approaches \cite{raileanu2021automatic, yarats2020image, nguyen2024symmetry},  we construct an equivariant policy through our addition of the equivariant ensembles, such that the value regularization is theoretically sound.

\section{Methodology}

Given that the optimal policy for a symmetric MDP is equivariant and the corresponding value function is invariant, one approach is to design neural network architectures for the actor and critic to be equivariant and invariant, respectively. However, this can be very challenging and highly dependent on the type of transformations applied. An alternative would be to augment the training data with the equivariant transitions to enrich the transition data. However, the na\"ive explicit equivariant data augmentation for PO,
\begin{equation}
\resizebox{0.9\linewidth}{!}{
    $J_\text{PO}(\phi) = \mathbb{E}_{(s, a)\sim\tau^{\pi_{\phi_\text{old}}}}\left[\frac{1}{|G|}\sum_{g\in G}\frac{\pi_\phi(K_g[a]|L_g[s])}{\pi_{\phi_\text{old}}(a|s)}\mathrm{A}^{\pi_{\phi_\text{old}}}(s,a)\right]$,} 
\end{equation}
is not a sound estimate if $\pi_\phi$ is not equivariant, i.e., $\pi_\phi(K_g[a]|L_g[s]) \neq \pi_{\phi}(a|s)$ as was already noted by \cite{raileanu2021automatic} in the case of invariant data augmentation. 
Additionally, if $\pi_\phi$ is equivariant, the policy optimization objective reduces back to the original one in \eqref{eq:po_loss} as all elements of the summation are equal. Therefore, the question is, how can gradients of the PO objective be propagated to the actor for all transformations in $G$? We achieve this by constructing an equivariant policy that explicitly incorporates all transformations: Equivariant Ensembles.

\subsection{Equivariant Ensembles}
\label{sec:policy_ens}

By averaging the policy for each transformation, we create the policy ensemble as
\begin{equation}\label{eq:ens}
    \Bar{\pi}_\phi(\cdot|s) = \frac{1}{|G|}\sum_{g\in G} P_g^{-1}[\pi_\phi(\cdot|L_g[s])].
\end{equation}
While the normal policy is likely not equivariant, i.e., $P_h[\pi(\cdot|s)] \not\equiv \pi(\cdot|L_h[s])$ for all $h\in G$, the policy ensemble is equivariant by construction, i.e.,
\begin{equation}\label{eq:ens_equiv}
    P_h[\Bar{\pi}_\phi(\cdot|s)] = \Bar{\pi}_\phi(\cdot|L_h[s]), \quad\forall h\in G.
\end{equation}

\begin{proof}
    Applying the transformation $P_h$ to the ensemble policy in \eqref{eq:ens} yields
\begin{align}
    P_h[\Bar{\pi}_\phi(\cdot|s)] =& P_h\left[\frac{1}{|G|}\sum_{g\in G} P_g^{-1}[\pi_\phi(\cdot|L_g[s])]\right] \\
    =& \frac{1}{|G|}\sum_{g\in G} P_h[P_g^{-1}[\pi_\phi(\cdot|L_g[s])]],
\end{align}
interchanging it with the summation and constant terms since $P_h$ is a linear operator. Recall that the composition of the transformations $P_h$ and $P_g^{-1}$ is also in $G$. Therefore, for each $g\in G$ there exists an $g' = h \circ g^{-1}$ (with $g = g'^{-1} \circ h$) in $G$, allowing the reindexation of the sum as 
\begin{equation}
    P_h[\Bar{\pi}_\phi(\cdot|s)] = \frac{1}{|G|}\sum_{g'\in G} P_{g'}[\pi_\phi(\cdot|L_{g'^{-1} \circ h}[s])]
\end{equation}
Since $L_{g'^{-1} \circ h}[s] = L_{g'}^{-1}[L_{h}[s]]$, it can be rewritten as 
\begin{equation}
    P_h[\Bar{\pi}_\phi(\cdot|s)] = \frac{1}{|G|}\sum_{g'\in G} P_{g'}[\pi_\phi(\cdot|L_{g'}^{-1}[L_{h}[s]])
\end{equation}
Since the inverse of $g'$ is part of $G$, the summation is simply rearranged when iterating over $g = g'^{-1}$, yielding
\begin{align}
    P_h[\Bar{\pi}_\phi(\cdot|s)] =& \frac{1}{|G|}\sum_{g\in G} P_{g}^{-1}[\pi_\phi(\cdot|L_{g}[L_{h}[s]])]\\
    =& \Bar{\pi}_\phi(\cdot|L_h[s]), 
\end{align}
concluding the proof of equivariance of the ensemble policy.
\end{proof}
Using the policy ensemble as behavior policy, the PO objective can be written as
\begin{equation}\label{eq:po_loss_new}
    J_\text{PO}(\phi) = \mathbb{E}_{(s, a)\sim\tau^{\Bar{\pi}_{\phi_\text{old}}}}\left[\frac{\Bar{\pi}_\phi(a|s)}{\Bar{\pi}_{\phi_\text{old}}(a|s)}\mathrm{A}^{\Bar{\pi}_{\phi_\text{old}}}(s,a)\right],
\end{equation}
which allows the gradient to propagate to the parameters $\phi$ for all transformations as they are incorporated in the ensemble policy in \eqref{eq:ens}. Similarly, an ensemble critic can be created as
\begin{equation}\label{eq:value_ens}
    \Bar{\mathrm{V}}^{\Bar{\pi}}_\theta(s) = \frac{1}{|G|}\sum_{g\in G}V^{\Bar{\pi}}_\theta(L_g[s])
\end{equation}
to leverage the same effect during critic training. The value ensemble corresponding to the equivariant ensemble policy is invariant by construction, i.e.,
\begin{equation}
    \Bar{\mathrm{V}}^{\Bar{\pi}}_\theta(L_g[s]) = \Bar{\mathrm{V}}^{\Bar{\pi}}_\theta(s), \quad \forall g \in G
\end{equation}

\begin{proof}
The invariance of the ensemble critic can be shown through the following derivation:
    \begin{align}
        \Bar{\mathrm{V}}^{\Bar{\pi}}_\theta(L_g[s]) &= \frac{1}{|G|}\sum_{g'\in G}V^{\Bar{\pi}}_\theta(L_{g'}[L_g[s]]) & | h = g'\circ g \\
        &= \frac{1}{|G|}\sum_{h\in G}V^{\Bar{\pi}}_\theta(L_h[s]) & | g = h\\
        &= \Bar{\mathrm{V}}^{\Bar{\pi}}_\theta(s)
    \end{align}
concluding the proof.
\end{proof}

Using the ensemble actor and critic thus allows the PO gradients to propagate through the networks for each possible transformation at the same time, which should improve training efficiency and generalization, as the agent observes a wider variety of scenarios at the same time. 

\subsection{Equivariant Regularization}

Even though the ensemble policy is used as behavior policy and the actor is thus trained for all transformations simultaneously, the individual policy for one transformation is not explicitly trained to be equivariant. However, explicitly training the actor and critic to be equivariant and invariant could give some inductive bias that could further accelerate training. Therefore, we add a regularization loss for the actor and critic that penalizes the difference to the ensemble.

For the actor, the regularization loss is defined as 
\begin{equation}
    \mathrm{L}_\pi(\phi) = \mathbb{E}_{s\sim\tau}\left[\frac{1}{|G|} \sum_{\forall g \in G} \mathrm{D}\left( \pi_\phi(\cdot | L_g[s]) \big|\big| \Bar{\pi}_\phi(\cdot | L_{g}[s])] \right)  \right],
\end{equation}
where $\mathrm{D}$ is a divergence measure such as the Kullback-Leibler (KL) divergence, $\mathrm{D}_\text{KL}$. This loss pushes the actor to output the equivariant distribution given each individual transformation. Similarly, the critic regularization can be formulated as
\begin{equation}
    \mathrm{L}_V(\theta) = \mathbb{E}_{s\sim\tau}\left[ \frac{1}{|G|} \sum_{\forall g \in G} \left(\mathrm{V}_\theta^{\Bar{\pi}}(L_g[s]) - \Bar{\mathrm{V}}^{\Bar{\pi}}_\theta(s)\right)^2 \right]
\end{equation}
which is a mean square error on the difference between the value estimates of all transformations to the ensemble critic. 

\section{Map-based Path Planning Case Study}

\begin{figure}
\vspace{4pt}
\centering
\begin{minipage}{0.49\columnwidth}
\includegraphics[width=\textwidth]{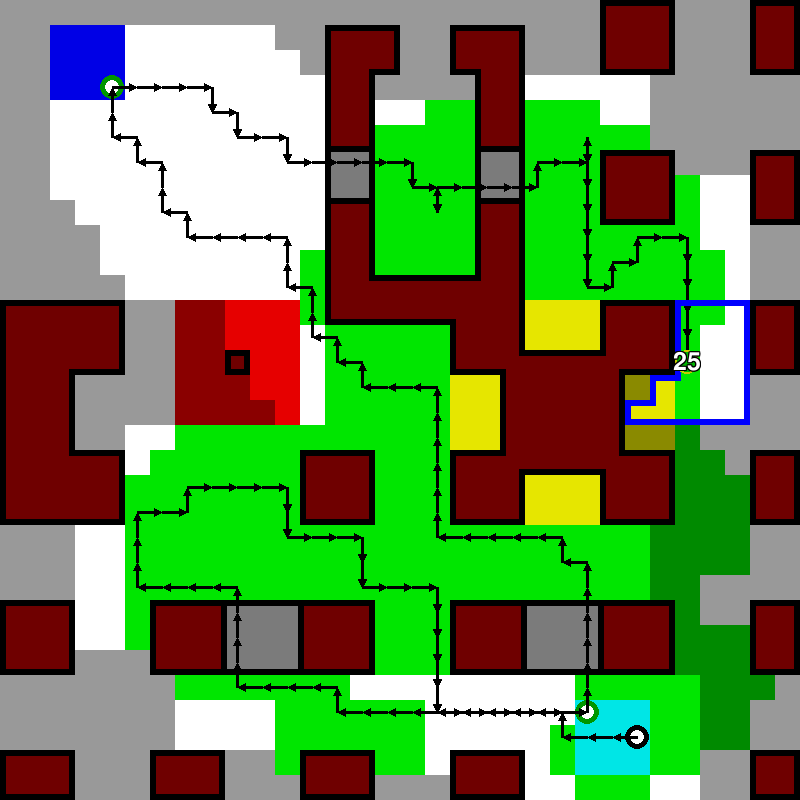}
\end{minipage}%
\hfill%
\begin{minipage}{0.5\columnwidth}
\small
\renewcommand{\arraystretch}{1.0}
\begin{tabular*}{\textwidth}{cl}
\toprule[1.5pt]
 & Description\\
\midrule
\includegraphics[align=c,height=.3cm]{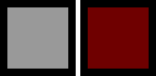} & Low/High buildings\\
\includegraphics[align=c,height=.3cm]{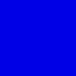} & Landing zone\\
\includegraphics[align=c,height=.3cm]{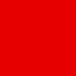} & No-fly zone (NFZ)\\
\includegraphics[align=c,height=.3cm]{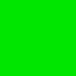} & Target zone \\
\includegraphics[align=c,height=.3cm]{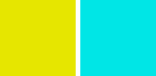} & Target + NFZ/Landing\\
\includegraphics[align=c,height=.3cm]{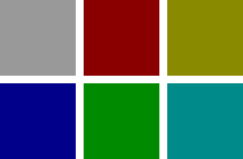} & Cells not covered\\
\includegraphics[align=c,height=.3cm]{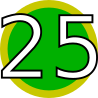} & Agent with battery\\
\includegraphics[align=c,height=.3cm]{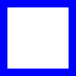} & Field of view (FoV)\\
\includegraphics[align=c,width=.5cm]{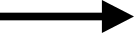} & Trajectory\\
\includegraphics[align=c,height=.3cm]{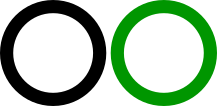} & Take-off/Charging\\
\bottomrule[1.5pt]
\end{tabular*}
\end{minipage}
\caption{Example state of a UAV in a coverage path planning grid-world problem on the left, showing the covered area, trajectory, and field of view, with a legend on the right.}
\label{fig:cpp}
\end{figure}

Map-based path planning is a widely spread problem for various mobile robots. In this paper, we consider the UAV coverage path planning (CPP) problem, as it is a challenging problem with rotational symmetries.

\subsection{Problem Formulation}

In the UAV CPP problem, a UAV equipped with a face-down camera is tasked to fly over designated target zones while avoiding obstacles and adhering to battery constraints. This paper considers the power-constrained CPP problem with recharge, defined in \cite{theile2023learning}. The agent is a UAV moving in a grid world, as shown in Figure~\ref{fig:cpp}. As the legend describes, the environment consists of obstacles, landing zones, and no-fly zones (NFZs). The obstacles can be low such that the agent can fly over or high so it cannot. The environment contains target zones, which the agent is supposed to cover. The agent is a quadcopter-like UAV that can move north, east, south, or west and take off, land, or recharge in landing zones. It has an onboard battery, indicated by the number of steps the agent can do before it runs out. The UAV is equipped with a camera whose rectangular field of view (FoV) is obstructed by low and high obstacles. The objective is to find the shortest trajectory such that each cell of the target zone is in the FoV at least once, i.e., each cell is covered. After covering all target zones, the task is defined as solved after the agent lands in one of the landing zones.

General CPP problems are proven to be NP-hard \cite{arkin2000approximation}, with the power-constrained UAV CPP problem with recharge problem adding to the challenge in two ways. First, since the UAV is equipped with a camera, its FoV is larger than its occupancy footprint, meaning it can cover cells by visiting nearby cells. Therefore, formulating it as an optimization problem is more complex as there is more than one position of the UAV from which a target zone can be covered. Second, the power constraint and the ability to recharge in designated areas add significant complexity to the problem. The optimal trajectory hinges on efficiently dividing the target zone for multiple flight segments interleaved with charging periods. Additionally, the allocation of selected target zones for the return journeys exacerbates the challenge. Overall, this problem is a complex long-horizon problem that we chose to address with deep reinforcement learning.

\subsection{Reinforcement Learning for CPP}

\begin{figure}
\vspace{4pt}
    \centering
    \includegraphics[width=0.75\linewidth]{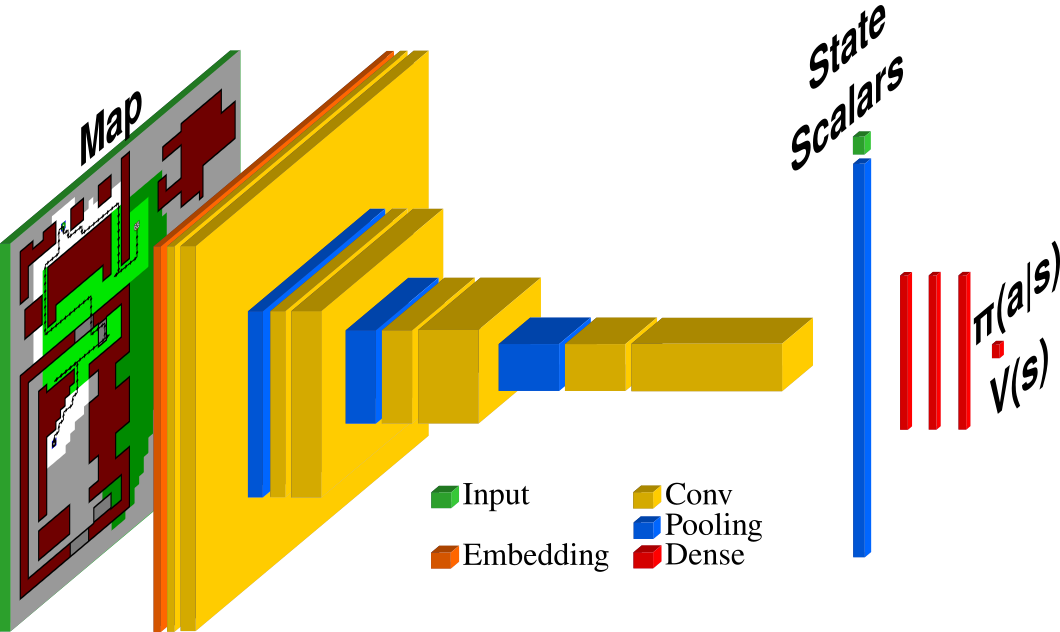}
    \caption{Neural network architecture where the position is a one-hot representation in the map, and the battery and landing state are fed in as state scalars.}
    \label{fig:nn_arch}
\end{figure}

As introduced in \cite{theile2020uav}, we address the UAV CPP problem with RL using map-based observations of the environment. The MDP of the problem contains the state space
\begin{equation}\label{eq:state_space}
    \mathcal{S} = 
    \underbrace{\mathbb{B}^{m\times m \times 3}}_{\substack{\text{Environment}\\ \text{Map}}}\times 
    \underbrace{\mathbb{B}^{m\times m}}_{\substack{\text{Target}\\ \text{Map}}}\times 
    \underbrace{\mathbb{N}^2}_{\text{Position}}\times
    \underbrace{\mathbb{N}}_{\substack{\text{Battery}\\ \text{Level}}}\times
    \underbrace{\mathbb{B}}_{{\text{Landed}}},
\end{equation}
given the number of cells in the environment as $m\times m$. The action space is defined as a discrete set of actions 
\begin{equation}\label{eq:action_space}
\mathcal{A} = \{\textit{east}, \textit{north}, \textit{west}, \textit{south}, \textit{take off}, \textit{land}, \textit{charge}\}.
\end{equation}
The neural network to process the state input and create the action of value output is shown in Figure~\ref{fig:nn_arch}. For the details of the transition function $\mathrm{P}$, the interested reader is directed to \cite{theile2023learning}. The reward function yields the step-wise reward
\begin{equation}
    r_t = r_c (|\mathcal{C}_{t}| - |\mathcal{C}_{t+1}|) - r_m,
\end{equation}
which gives a reward for the difference in the number of covered target cells $|\mathcal{C}_{t}| - |\mathcal{C}_{t+1}|$ scaled by $r_c$ after performing the action and a flat penalty $r_m$ for every step. The first part of the reward incentivizes the agent to cover as much of the target zones as possible, while the second part directs the agent to finish as early as possible to stop incurring penalties. As in \cite{theile2023learning}, we schedule the discount factor $\gamma$ such that the agent first learns to solve the task using a lower $\gamma$ and then optimizes path length with a $\gamma \rightarrow 1$ where the cumulative movement penalty becomes dominant. In this paper, however, we do not schedule the discount factor based on training steps, as it differs based on the algorithm. Instead, we schedule it based on performance, such that the discount factor increases slightly every time the agent successfully solves a scenario. We further adopt the action masking approach from \cite{theile2023learning}, which masks out actions that would lead to a collision or drive the agent too far away from a landing zone.

\subsection{Equivariances in the CPP Problem}

The equivariances utilized in this paper are shown in Figure~\ref{fig:transforms}. We only focus on the four rotations as $L_g$ and leave horizontal and vertical flips to future work. When rotating the input map, the three actions \textit{take off}, \textit{land}, and \textit{charge} remain unaffected, while the four directional actions permute depending on the rotation applied. 
\section{Experiment}
\label{sec:results}
\subsection{Setup}

\begin{figure}
\vspace{4pt}
    \centering
    \begin{subfigure}{0.9\columnwidth}
        \resizebox{0.98\linewidth}{!}{
        \begin{tikzpicture}
            \node[rectangle,draw=black,line width=2pt,inner sep=0pt] (id1) at (0,0) {\includegraphics[width=3.2cm]{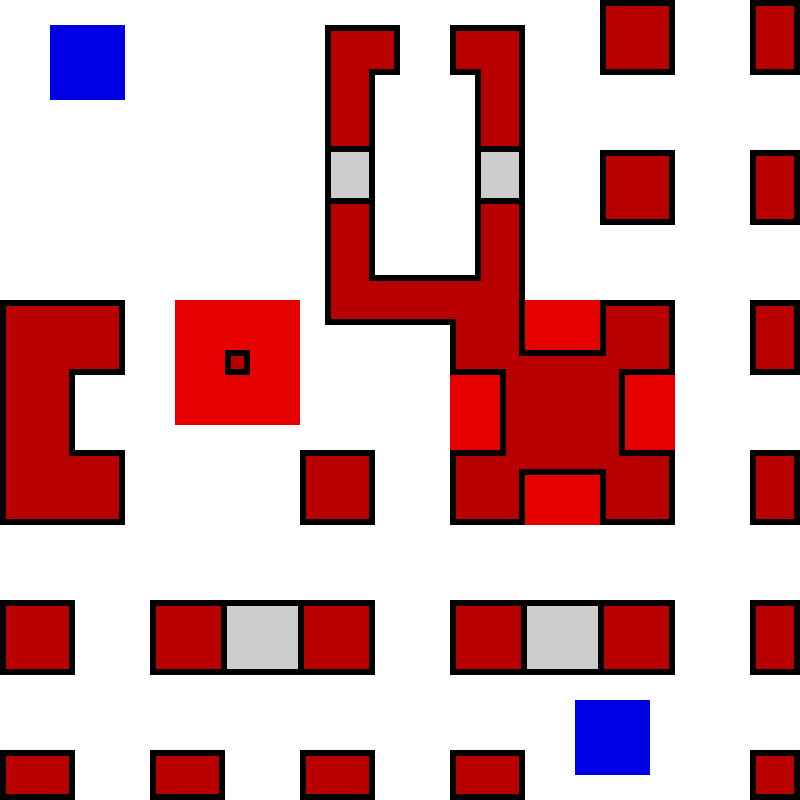}};
            \node[rectangle,draw=black,line width=2pt,inner sep=0pt] (id2) at (3.3,0) {\includegraphics[width=3.2cm]{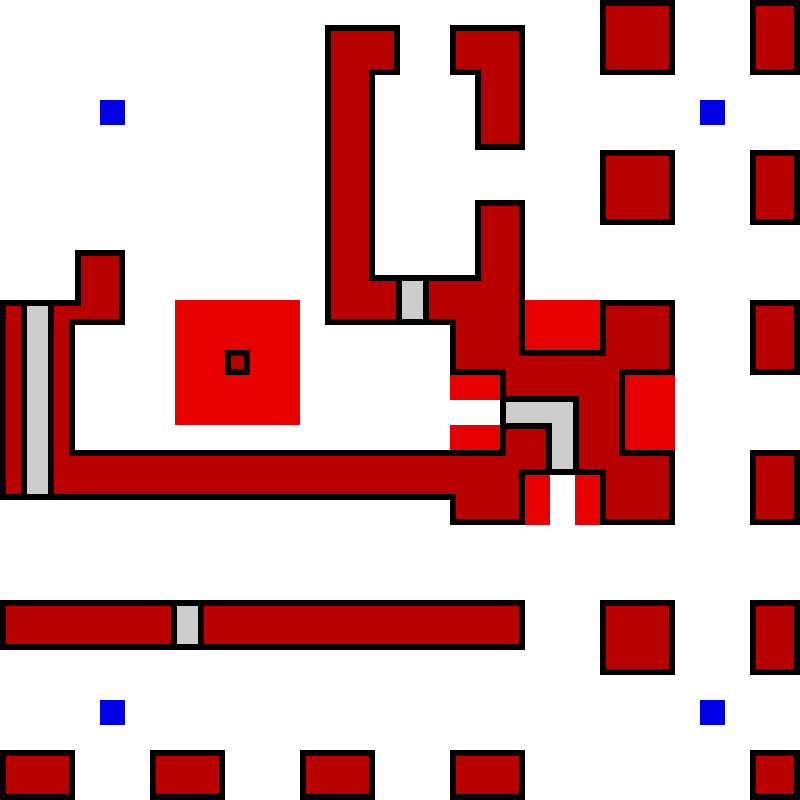}};
            \node[rectangle,draw=black,line width=2pt,inner sep=0pt] (id3) at (6.6,0) {\includegraphics[width=3.2cm]{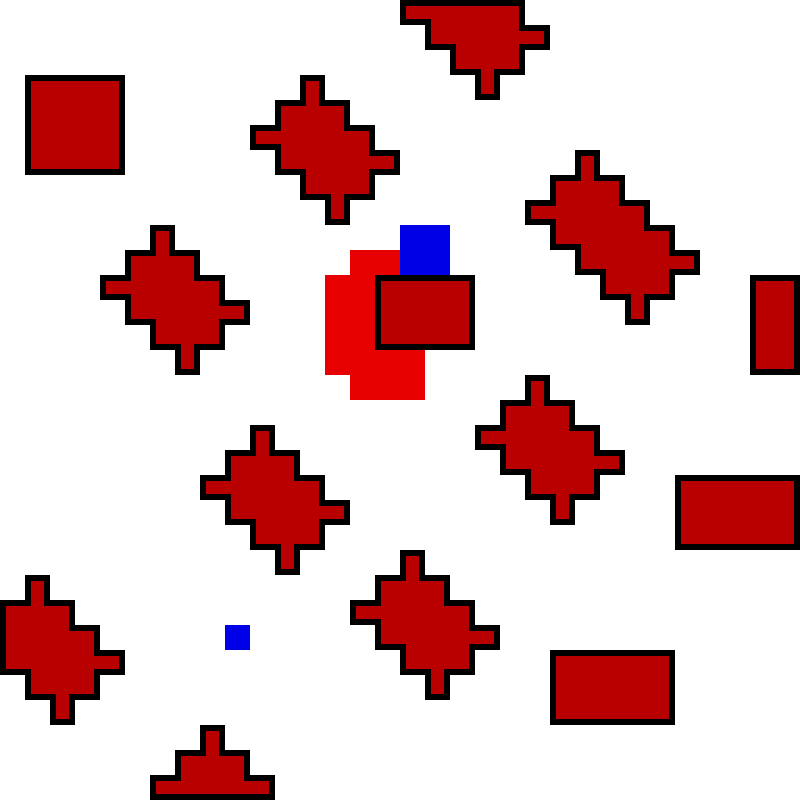}};
            \node[rectangle,draw=black,line width=2pt,inner sep=0pt] (id3) at (10.2,0) {\includegraphics[width=3.8cm]{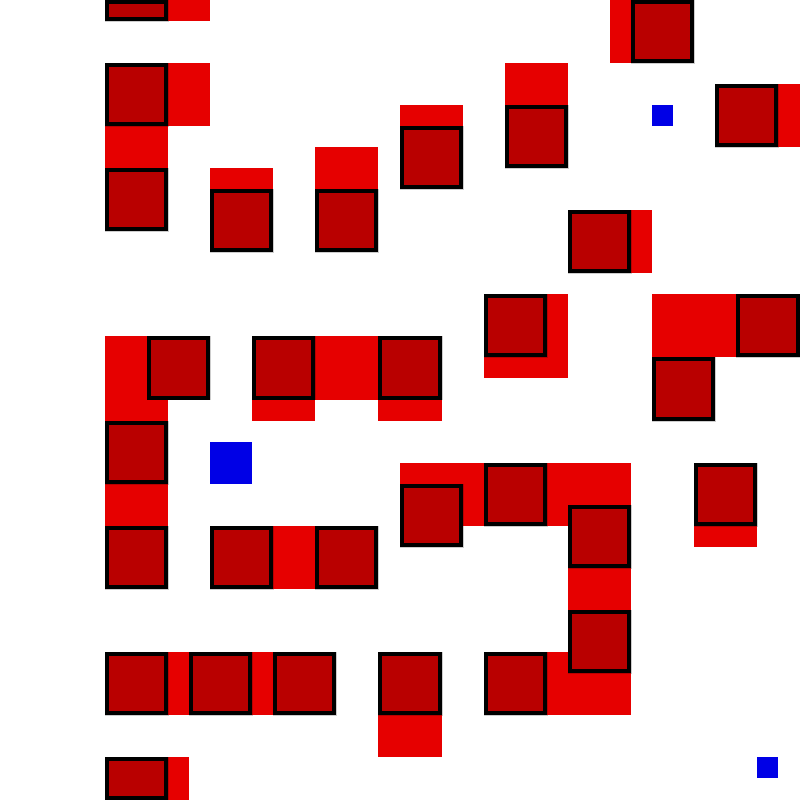}};
            \node[rectangle,draw=black,line width=2pt,inner sep=0pt] (id3) at (14.2,0) {\includegraphics[width=4cm]{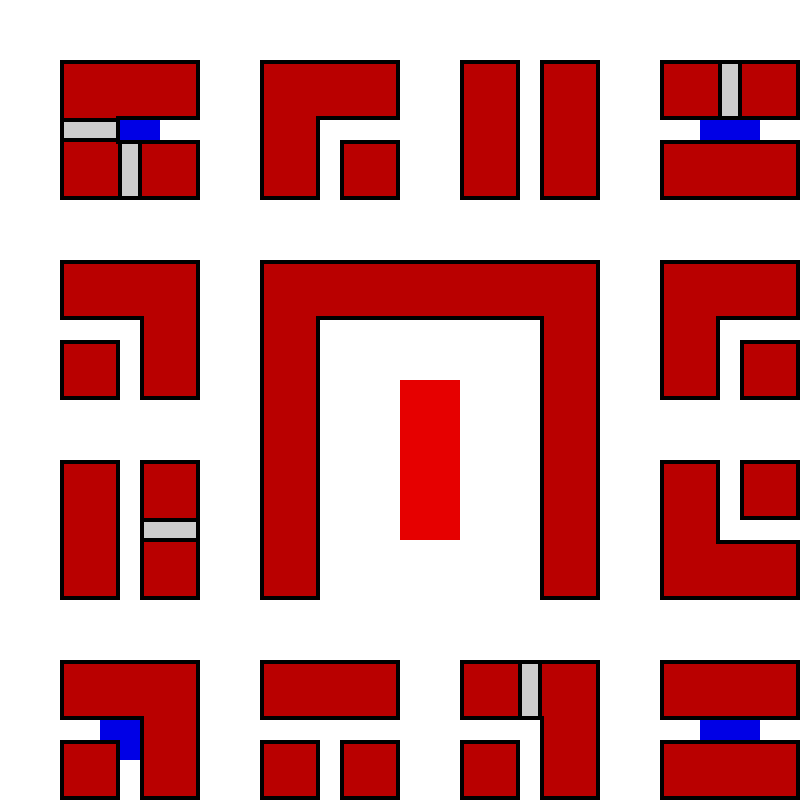}};
            \node[rectangle,draw=black,line width=2pt,inner sep=0pt] (id3) at (18.5,0) {\includegraphics[width=4.4cm]{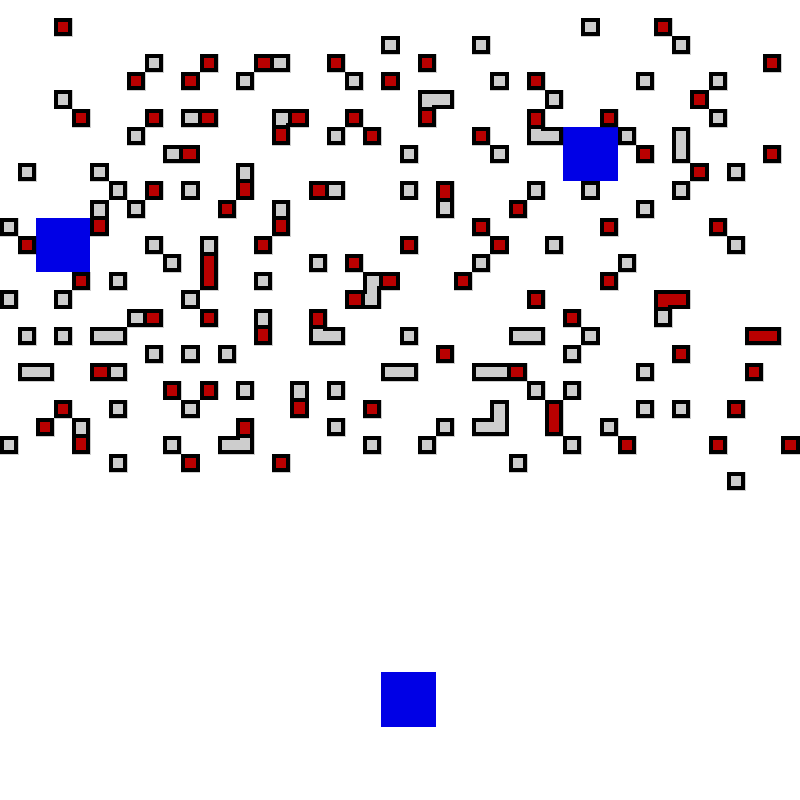}};
            
            \node[rectangle,draw=black,line width=2pt,inner sep=0pt] (id3) at (2.,-5.) {\includegraphics[width=5cm]{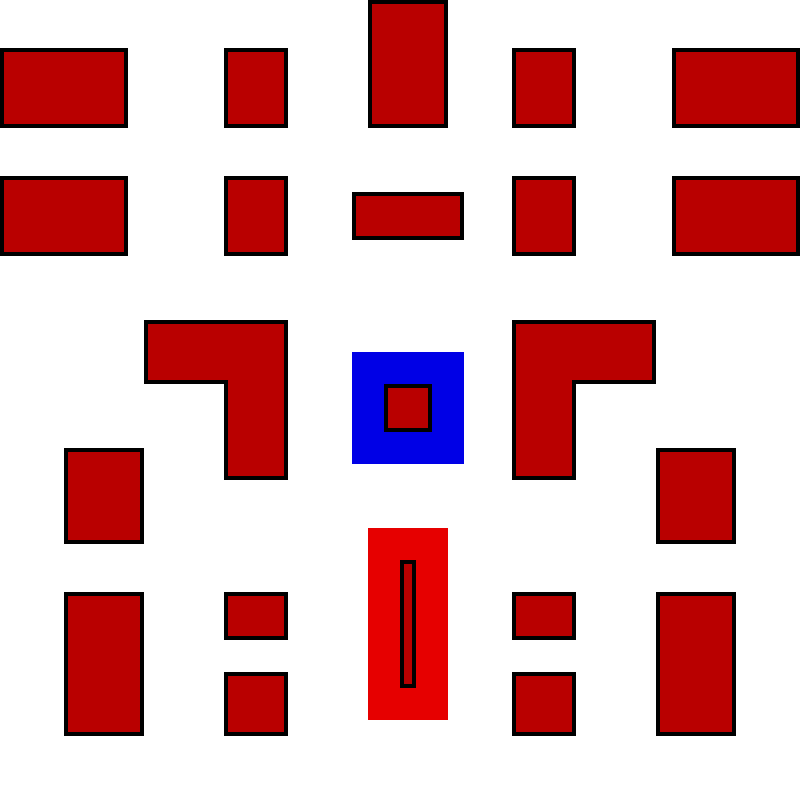}};
            \node[rectangle,draw=black,line width=2pt,inner sep=0pt] (id3) at (7.1,-5.) {\includegraphics[width=5cm]{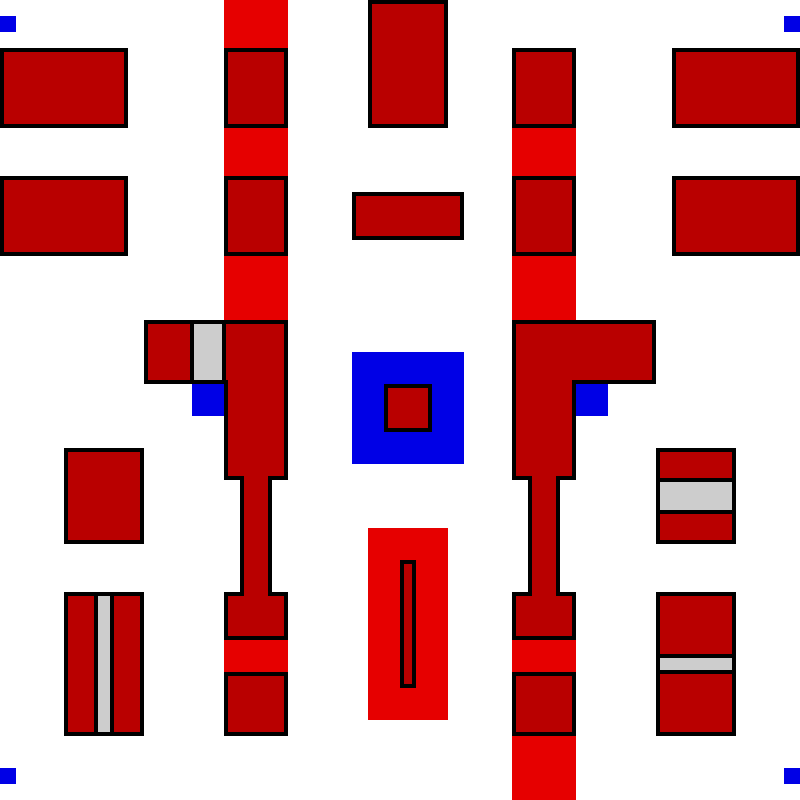}};
            \node[rectangle,draw=black,line width=2pt,inner sep=0pt] (id3) at (12.2,-5.) {\includegraphics[width=5cm]{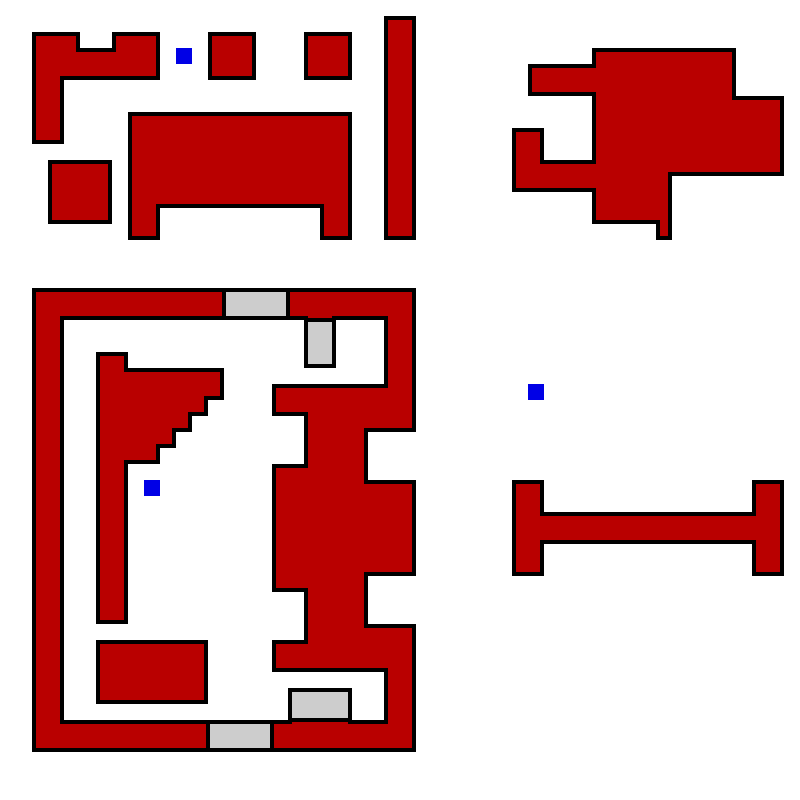}};
            \node[rectangle,draw=black,line width=2pt,inner sep=0pt] (id3) at (17.3,-5.) {\includegraphics[width=5cm]{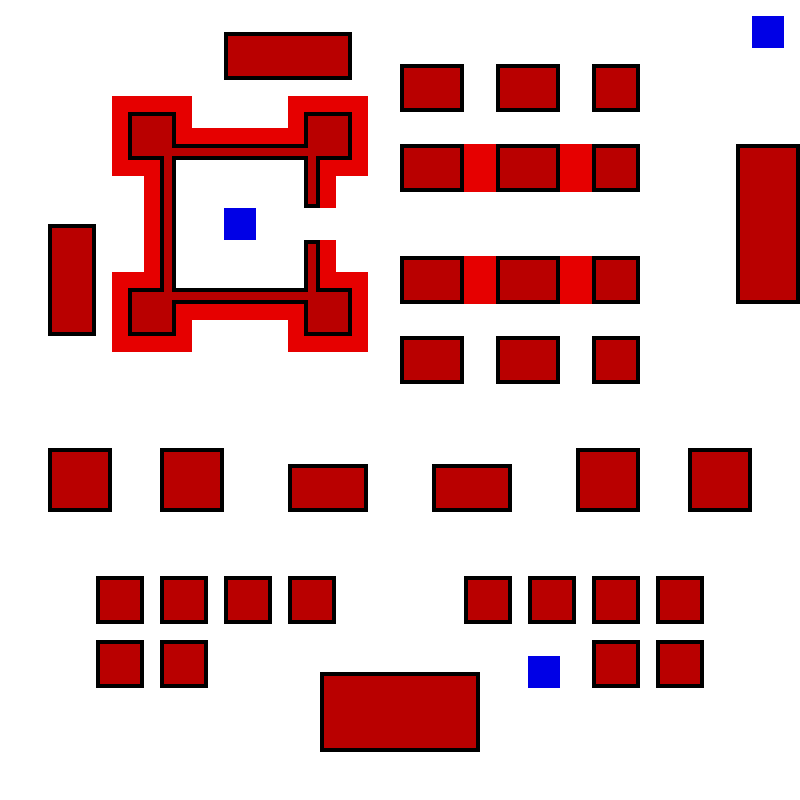}};
        \end{tikzpicture}
        }
    \caption{In-distribution maps.}
    \label{fig:id_maps}
    \end{subfigure}
    \begin{subfigure}{0.9\columnwidth}
        \resizebox{\linewidth}{!}{
        \begin{tikzpicture}
            \node[rectangle,draw=black,line width=2pt,inner sep=0pt] (id1) at (0,0) {\includegraphics[width=3.2cm]{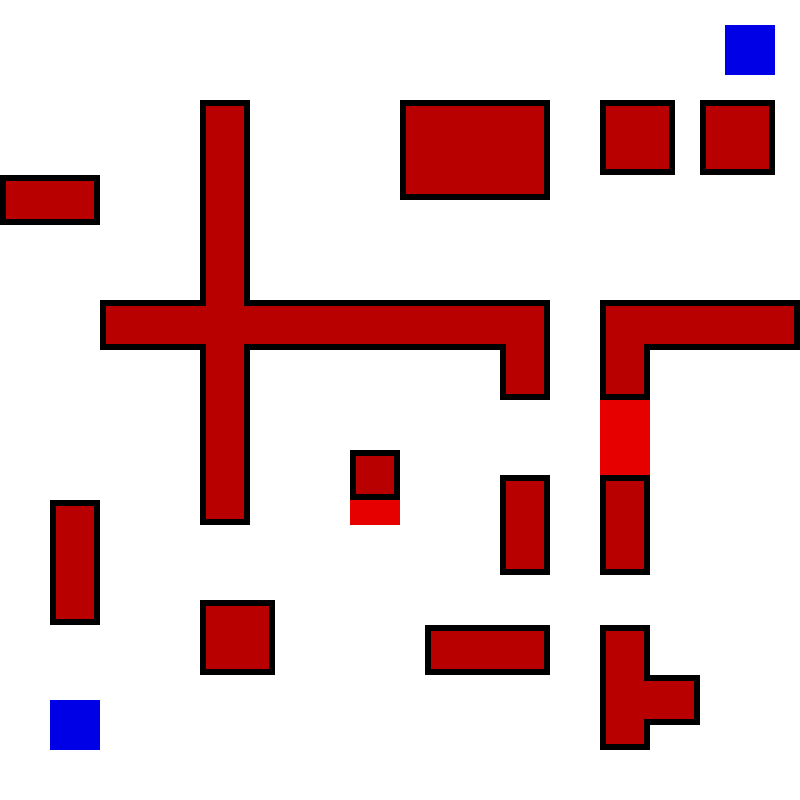}};
            \node[rectangle,draw=black,line width=2pt,inner sep=0pt] (id2) at (3.4,0) {\includegraphics[width=3.4cm]{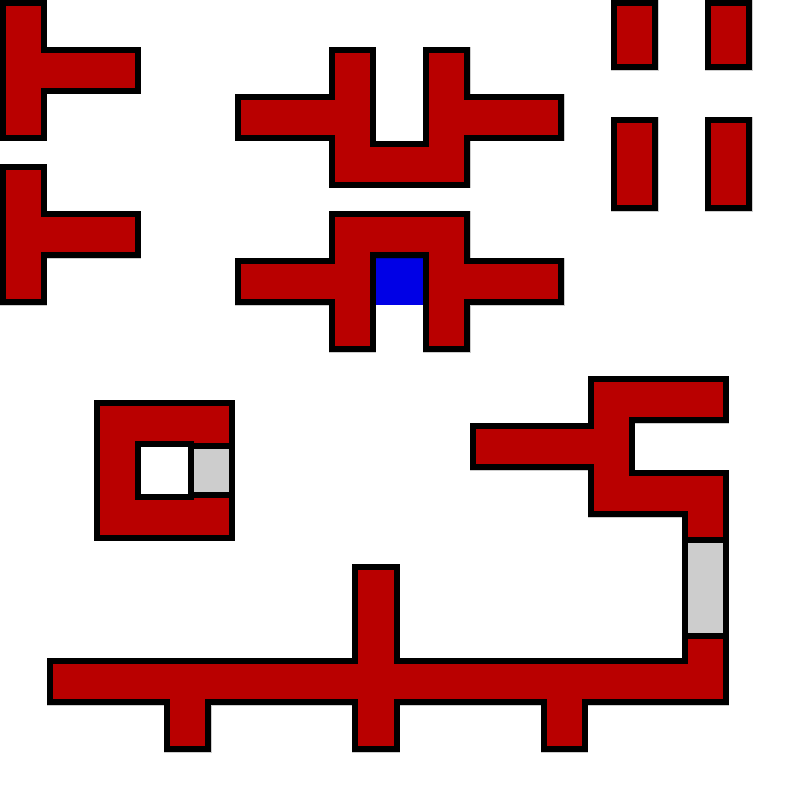}};
            \node[rectangle,draw=black,line width=2pt,inner sep=0pt] (id3) at (7.0,0) {\includegraphics[width=3.6cm]{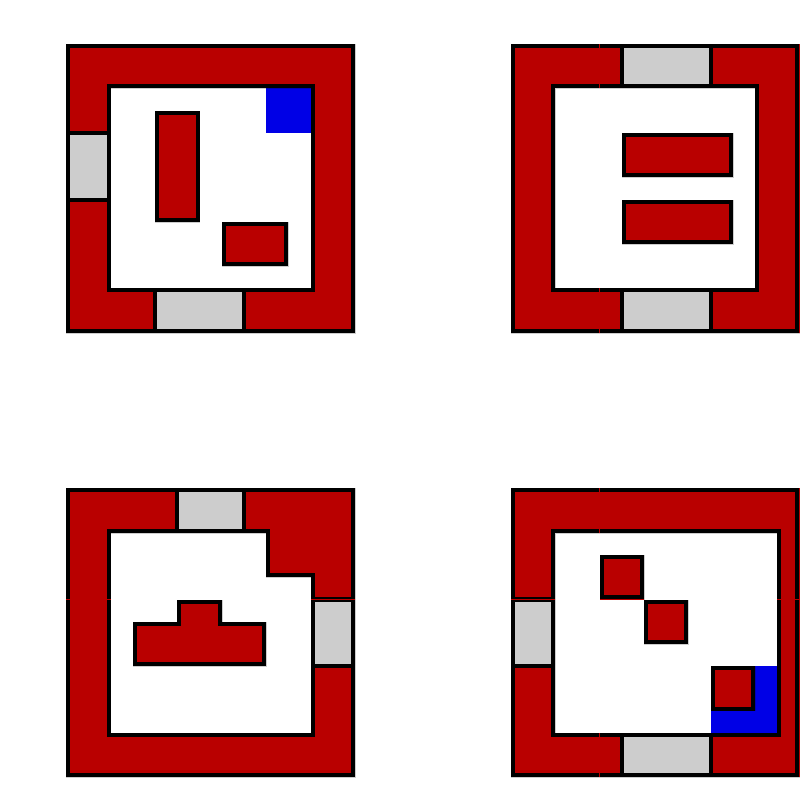}};
            \node[rectangle,draw=black,line width=2pt,inner sep=0pt] (id3) at (10.8,0) {\includegraphics[width=3.8cm]{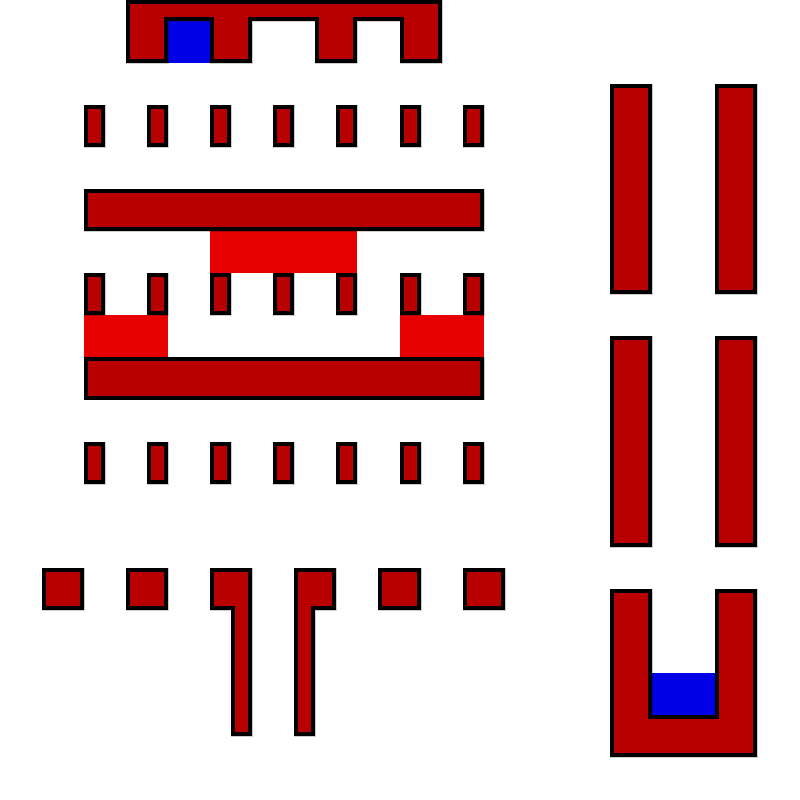}};
            \node[rectangle,draw=black,line width=2pt,inner sep=0pt] (id3) at (14.8,0) {\includegraphics[width=4cm]{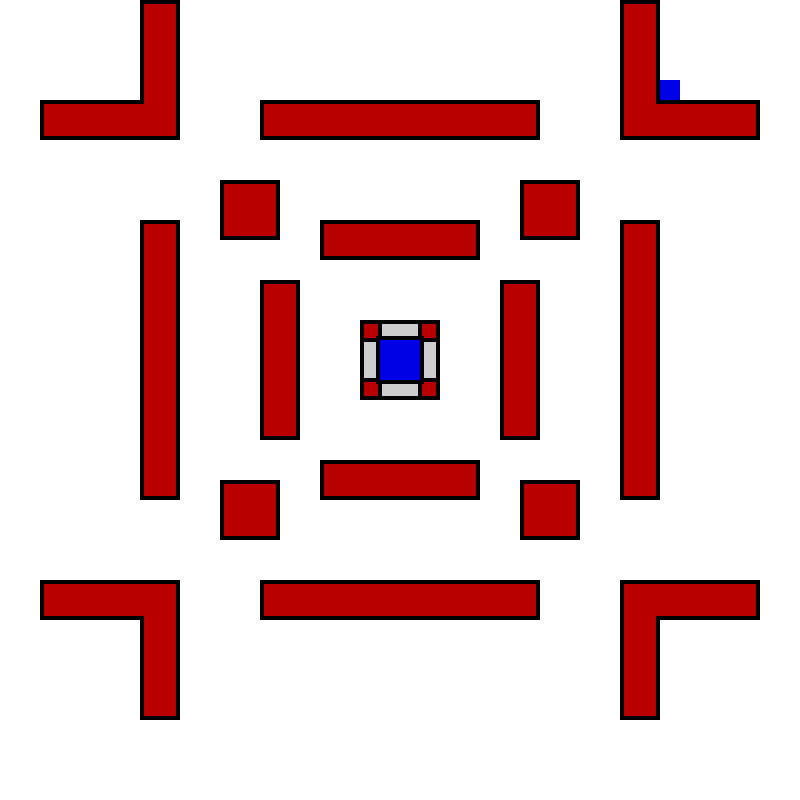}};
            \node[rectangle,draw=black,line width=2pt,inner sep=0pt] (id3) at (19.,0) {\includegraphics[width=4.2cm]{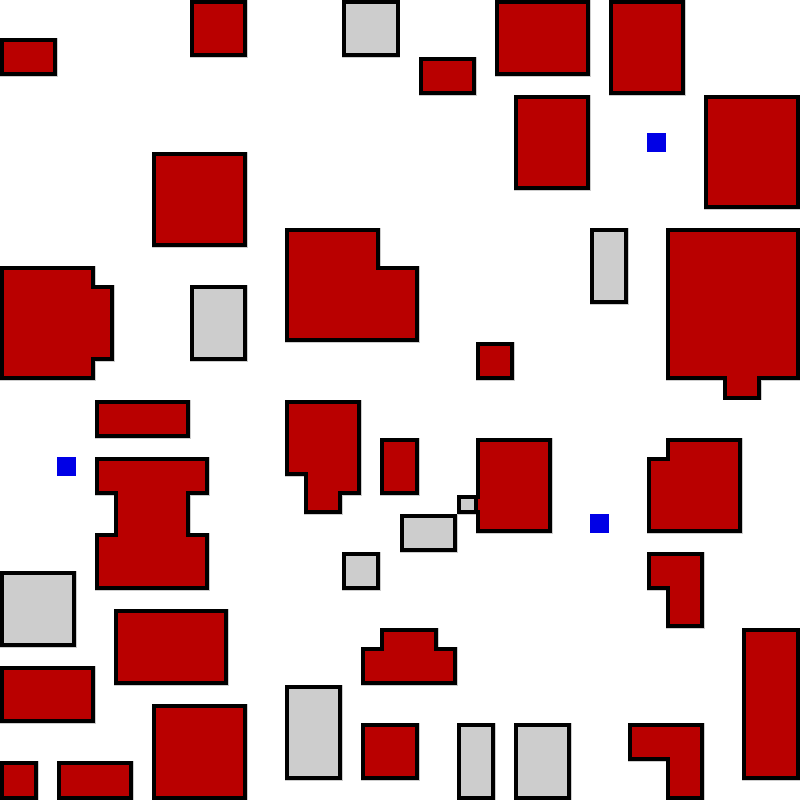}};
            
            \node[rectangle,draw=black,line width=2pt,inner sep=0pt] (id3) at (2.2,-4.8) {\includegraphics[width=4.4cm]{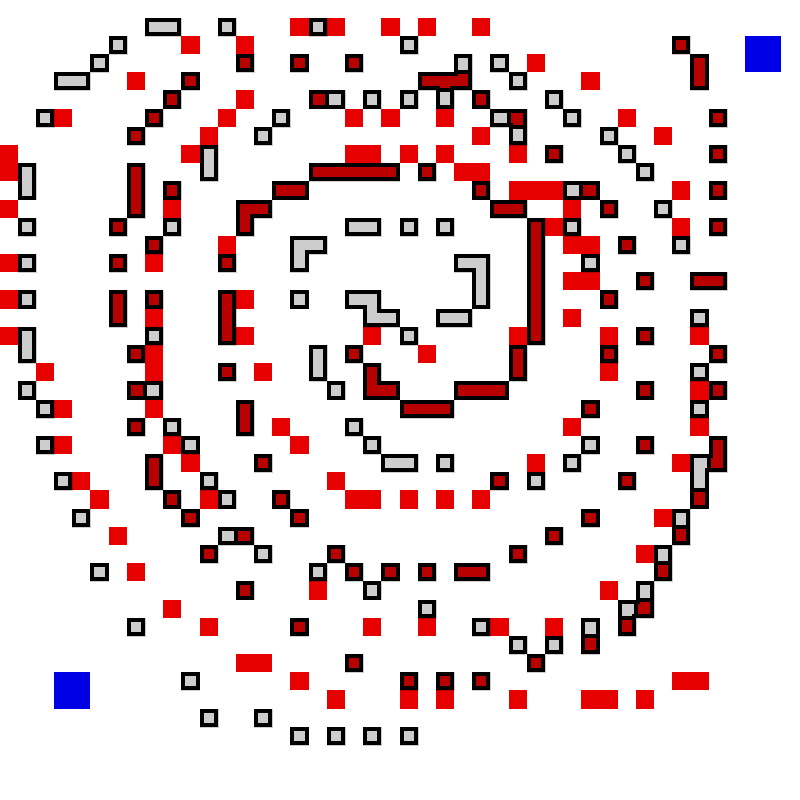}};
            \node[rectangle,draw=black,line width=2pt,inner sep=0pt] (id3) at (6.8,-4.8) {\includegraphics[width=4.6cm]{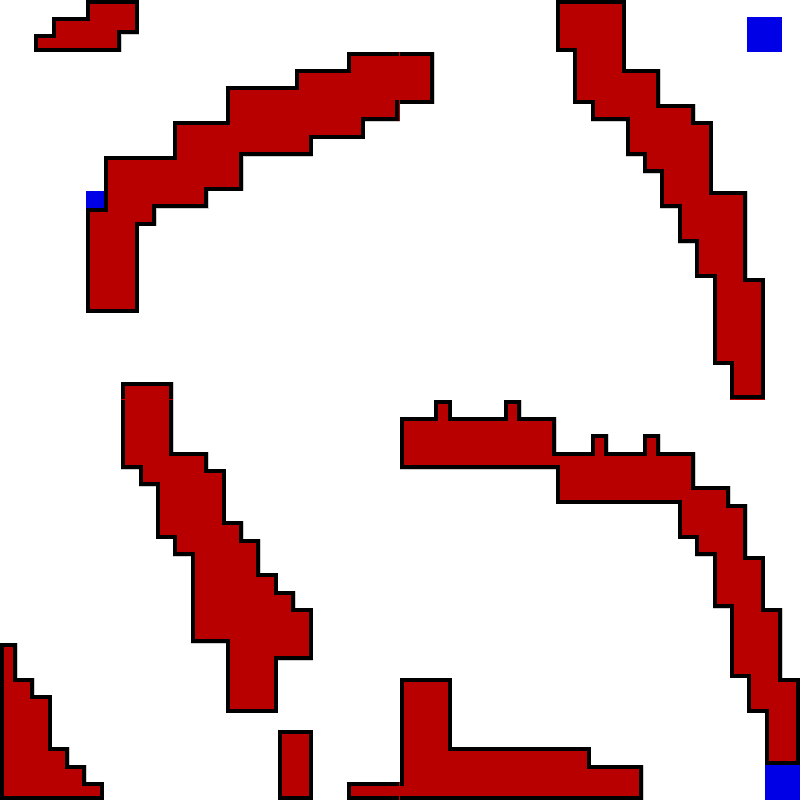}};
            \node[rectangle,draw=black,line width=2pt,inner sep=0pt] (id3) at (11.6,-4.8) {\includegraphics[width=4.8cm]{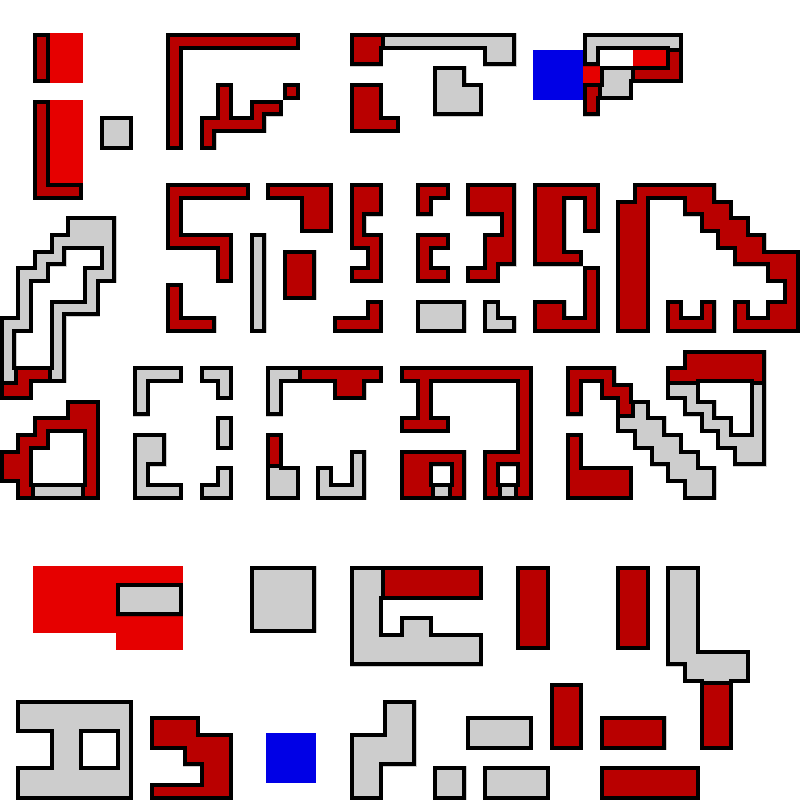}};
            \node[rectangle,draw=black,line width=2pt,inner sep=0pt] (id3) at (16.6,-4.8) {\includegraphics[width=5cm]{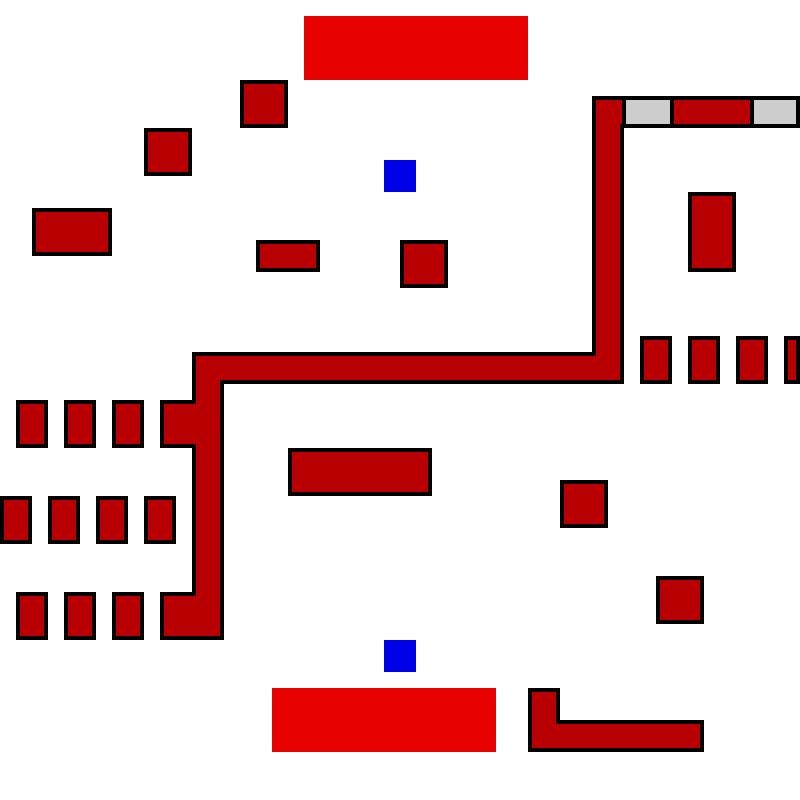}};
        \end{tikzpicture}
        }
    \caption{Out-of-distribution maps.}
    \label{fig:ood_maps}
    \end{subfigure}
    \caption{All maps used during training and testing.}
    \label{fig:all_maps}
\end{figure}

To study the effect of equivariant ensembles and regularization, we trained agents on 10 maps with varying sizes from $32\times 32$ to $50\times 50$ shown in Figure~\ref{fig:id_maps}. In each scenario encountered during training, one of the maps is randomly chosen, and a random target zone is generated. Since the range of possible target zones is vast, it is unlikely that an agent will ever encounter the same scenario between two episodes. Each agent is trained using PPO~\cite{schulman2017proximal} for 100M interaction steps with the environment with rollouts of 40K steps. 

We trained five different agent configurations: 
\begin{enumerate}
    \item \textit{Baseline}: Normal agent trained on 10 maps.
    \item \textit{Augment}: Agent trained on the 10 maps in all four rotations, yielding 40 training maps.
    \item \textit{Ensemble}: Agent with ensemble actor and critic.
    \item \textit{Regularized}: Agent with actor and critic regularization only.
    \item \textit{Ensemble + Regularized (Ens.+Reg.)}: Agent with ensemble actor and critic and regularization.
\end{enumerate}
The Augment agent was trained to show if the positive effects simply result from a broader set of maps trained on. Note that the Regularization agent regularizes the critic towards invariance even though the actor is not necessarily equivariant. For each agent configuration, three agents were trained, and their results were aggregated. The total training time for the Baseline and Augment agents is 24 hours and for the others, 60 hours on an Nvidia A100 GPU. The increase in training time stems from the added computation for the forward and backward passes for all transformations.

As metrics for comparison, we use
\begin{enumerate}
    \item \textit{Coverage ratio (CR)}: The ratio of covered cells when a timeout is reached, or the episode ended successfully, in which case CR=1.
    \item \textit{Task solved}: The share of scenarios that were solved, i.e., full coverage and landed, before a timeout (1500 steps) is reached.
    \item \textit{Episode steps}: The steps in the episode, which are 1500 if the timeout is reached, or the step of the landing action after the task is solved. The agent is supposed to minimize this metric.
    \item \textit{Relative deviation (RD)}: Comparing the steps needed to solve an episode with a heuristic (\cite{theile2023learning}), showing the difference. Lower is better.
\end{enumerate}
We will first show the training performance of the different agent configurations, followed by an analysis of their performance after training. In the last part, the effect of regularization on equivariance and invariance is visualized.

\subsection{Training Acceleration}

\begin{figure}
\vspace{5pt}
    \centering
    \begin{subfigure}{0.495\columnwidth}
        \includegraphics[width=\textwidth]{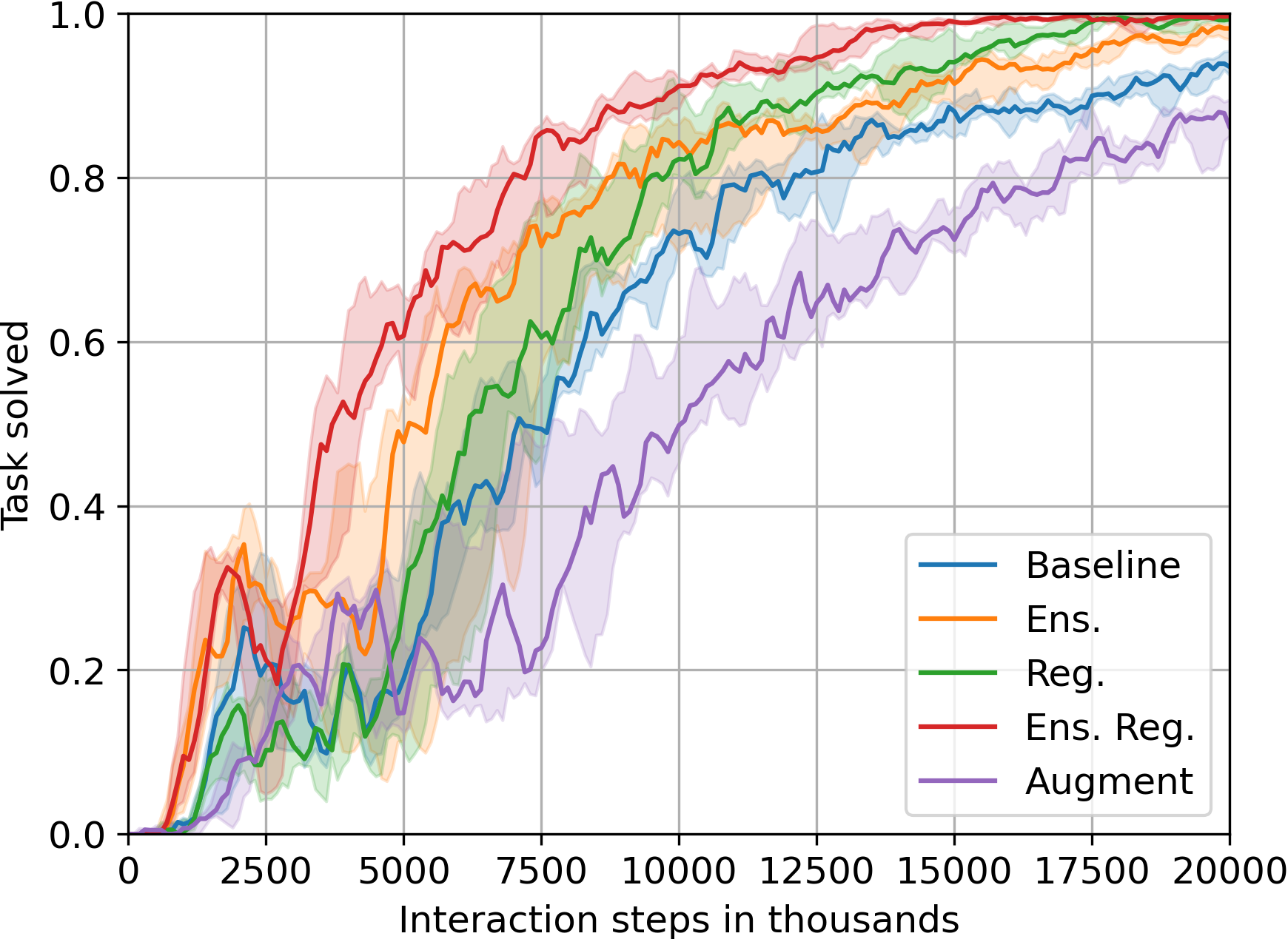}
        \caption{Task solved ratio}
        \label{fig:training:solved}
    \end{subfigure}%
    \begin{subfigure}{0.505\columnwidth}
        \includegraphics[width=\textwidth]{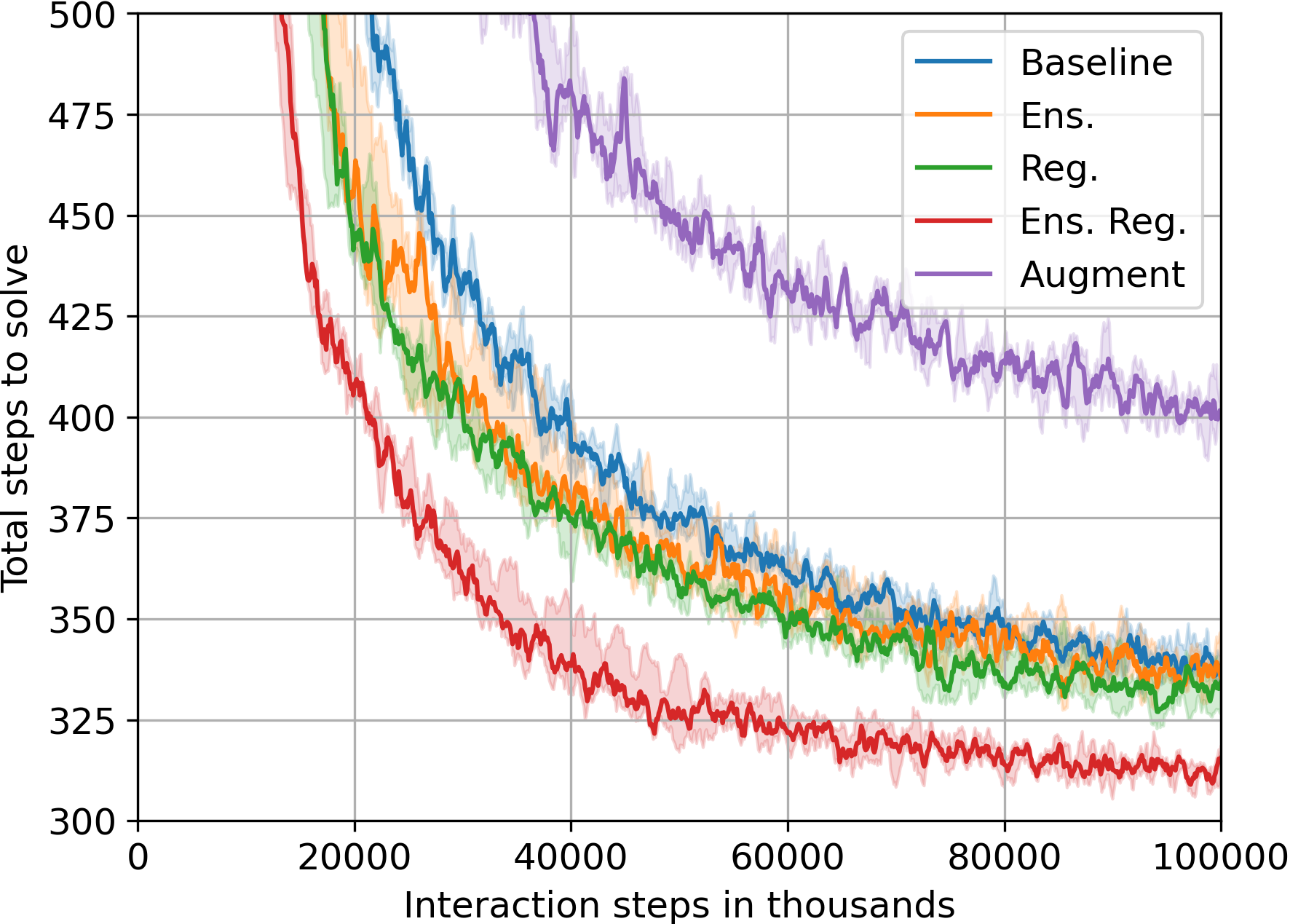}
        \caption{Steps to solve}
        \label{fig:training:steps}
    \end{subfigure}
    \caption{Training curves of all algorithms showing the task-solved ratio throughout the first 20M steps and the average steps to solve the scenarios for the full 100M training steps.}
    \label{fig:training}
\end{figure}

The training curves for the task-solved ratio and the steps to solve the episode are shown in Figure~\ref{fig:training}. The task-solved ratio in Figure~\ref{fig:training:solved}, which focuses on the first 20M interaction steps, shows how the ensemble and regularization accelerate the training individually and that their combination accelerates it even further. It further shows that the Augment agent takes longer to learn, as it interacts with 40 seemingly unrelated maps. In Figure~\ref{fig:training:steps}, it can be seen that ensemble and regularization individually accelerate training performance at first but converge to a similar final result as the baseline. The combination of ensemble and regularization, however, accelerates training and improves final performance significantly. On the other hand, the augmented agent struggles to find fast solutions, leading to slower and suboptimal learning performance. 

\subsection{Performance}

\begin{table*}
\vspace{4pt}
    \centering
\footnotesize
\setlength{\tabcolsep}{5pt}
\renewcommand{\arraystretch}{1.3}
  
\begin{tabular}{c| c|| c|c|c|c|c|c|c|c|c|c|c||c}
\multicolumn{2}{c||}{} & \multicolumn{2}{c|}{\multirow{1}{*}{Baseline}} & \multicolumn{2}{c|}{\multirow{1}{*}{Augment}} & \multicolumn{2}{c|}{\multirow{1}{*}{Ensemble}} &\multicolumn{2}{c|}{\multirow{1}{*}{Regularized}} &\multicolumn{2}{c||}{\multirow{1}{*}{Ens.+Reg.}} &\multicolumn{2}{c}{\multirow{1}{*}{Ens.+Reg.\raisebox{3pt}{\scriptsize 50M}}}   \\
\hline\hline
\multirow{2}{*}{In-Distribution (Fig~\ref{fig:id_maps})} 
% ==========
& Solved&\multicolumn2{c|}{$\mathbf{{1.00}^{0.00}}$} & \multicolumn2{c|}{$\mathbf{{1.00}^{0.00}}$} & \multicolumn2{c|}{$\mathbf{{1.00}^{0.00}}$} & \multicolumn2{c|}{$\mathbf{{1.00}^{0.00}}$} & \multicolumn2{c||}{$\mathbf{{1.00}^{0.00}}$} &\multicolumn2{c}{$\mathbf{{1.00}^{0.00}}$} \\\cline{2-14}
& RD&\multicolumn2{c|}{${-0.22}^{0.01}$} & \multicolumn2{c|}{${-0.15}^{0.01}$} & \multicolumn2{c|}{${-0.22}^{0.00}$} & \multicolumn2{c|}{${-0.24}^{0.01}$} & \multicolumn2{c||}{$\mathbf{{-0.28}^{0.00}}$} &\multicolumn2{c}{${-0.24}^{0.01}$}  
% ==========
\\ \hline \hline
                    
\multirow{2}{*}{In-Distribution Rotated} 
% ==========
& Solved&\multicolumn2{c|}{${0.58}^{0.03}$} & \multicolumn2{c|}{$\mathbf{{1.00}^{0.00}}$} & \multicolumn2{c|}{$\mathbf{{1.00}^{0.00}}$} & \multicolumn2{c|}{$\mathbf{{1.00}^{0.00}}$} & \multicolumn2{c||}{$\mathbf{{1.00}^{0.00}}$} &\multicolumn2{c}{$\mathbf{{1.00}^{0.00}}$} \\\cline{2-14}
& RD&\multicolumn2{c|}{${0.26}^{0.03}$} & \multicolumn2{c|}{${-0.14}^{0.01}$} & \multicolumn2{c|}{${-0.23}^{0.01}$} & \multicolumn2{c|}{${-0.19}^{0.00}$} & \multicolumn2{c||}{$\mathbf{{-0.28}^{0.00}}$} &\multicolumn2{c}{${-0.24}^{0.01}$} 
% ==========
\\ \hline \hline                  

\multirow{1}{*}{Out-of-Distribution (Fig~\ref{fig:ood_maps})} 
% ==========
& CR&\multicolumn2{c|}{${0.71}^{0.02}$} & \multicolumn2{c|}{${0.84}^{0.01}$} & \multicolumn2{c|}{$\mathbf{{0.86}^{0.02}}$} & \multicolumn2{c|}{${0.79}^{0.00}$} & \multicolumn2{c||}{${0.78}^{0.01}$} &\multicolumn2{c}{${0.80}^{0.00}$} 
% ==========
\\ \hline \hline
               
\end{tabular}
\caption{Quantitative comparison of the different agents tested on different sets of maps, showing the task-solved ratios and relative deviation (RD) to a heuristic for the in-distribution maps and the collection ratio (CR) for out-of-distribution maps. The Ens.+Reg.\raisebox{3pt}{\footnotesize 50M} agent is the performance of Ens.+Reg. after 50M interaction steps. The numbers show the median value of the three agents per configuration, with the maximum deviation in superscript.}
    \label{tab:comparison}
\end{table*}

This section compares the performance of each agent configuration on three different sets of maps: (1) the in-distribution maps that each agent encountered during training, (2) the in-distribution maps but rotated 90\textdegree, 180\textdegree, and 270\textdegree, and (3) 10 out-of-distribution maps shown in Figure~\ref{fig:ood_maps} that no agent saw during training. Table~\ref{tab:comparison} summarizes the results leading to the following observations. Each agent configuration consistently solves the in-distribution maps, with the Augment agent having the lowest RD performance (highest RD). As already seen in the training curves, the Ensemble and Regularized agents have similar performance compared with the baseline, and the combination has the best performance with a significant lead over the Baseline. 

For the rotated in-distribution maps, Augment, Ensemble, and Ens.+Reg. show very similar performance as for the non-rotated maps, showing that the equivariant ensemble is indeed equivariant. The Baseline cannot reliably solve these rotated maps. The Regularized agent shows a decreased performance for the rotated maps, which shows that it is not exactly equivariant, emphasizing the need for the ensemble when seeking an equivariant policy. 

Finally, the last row shows the average coverage ratio (CR) on the 10 out-of-distribution maps, as it is the best metric if the scenarios are not solved reliably. It can be seen that every agent configuration is better than the baseline, with the best-performing agent being the Ensemble, closely followed by the Augment agent. Regularization is hindering out-of-distribution performance, which we attribute to a specialization problem. The regularizing agents specialize on the in-distribution maps throughout training. This effect can be observed by inspecting the performance of the Ens.+Reg.\raisebox{3pt}{\footnotesize 50M} agent, which was only trained for 50M interaction steps. While its performance on in-distribution maps is lower than that of the fully trained counterpart, its CR value on the out-of-distribution maps is higher. To avoid this specialization effect in the future, the agents will probably need to be trained on a significantly more extensive set of maps or based on a procedural map generator.

To conclude, the study shows that the combination of equivariant ensembles and regularization significantly improves the performance for scenarios on the maps that were seen during training (note that the in-distribution \textit{scenarios} were still not seen during training as the optimal path strongly depends on the target zones that are generated randomly for each scenario). On out-of-distribution maps, just using the equivariant ensemble yields the highest performance as it appears to specialize less on the maps it has seen.

\subsection{Equivariance through Regularization}

\definecolor{blue}{HTML}{4C72B0}
\definecolor{orange}{HTML}{DD8452}
\definecolor{green}{HTML}{55A868}
\definecolor{red}{HTML}{C44E52}
\begin{table*}[]
    \centering
    \footnotesize
    \setlength{\tabcolsep}{2pt}
    \newcommand{\pheight}{1.2cm}
    \begin{tabu}{c|c||c|c|c|c|c}
    \multicolumn{2}{c||}{} & Baseline & Augment & Ensemble & Regularize & Ensemble + Regularize\\
    \hline\hline
       \multirow{2}[2]{*}[3.5mm]{\includegraphics[width=1.5cm]{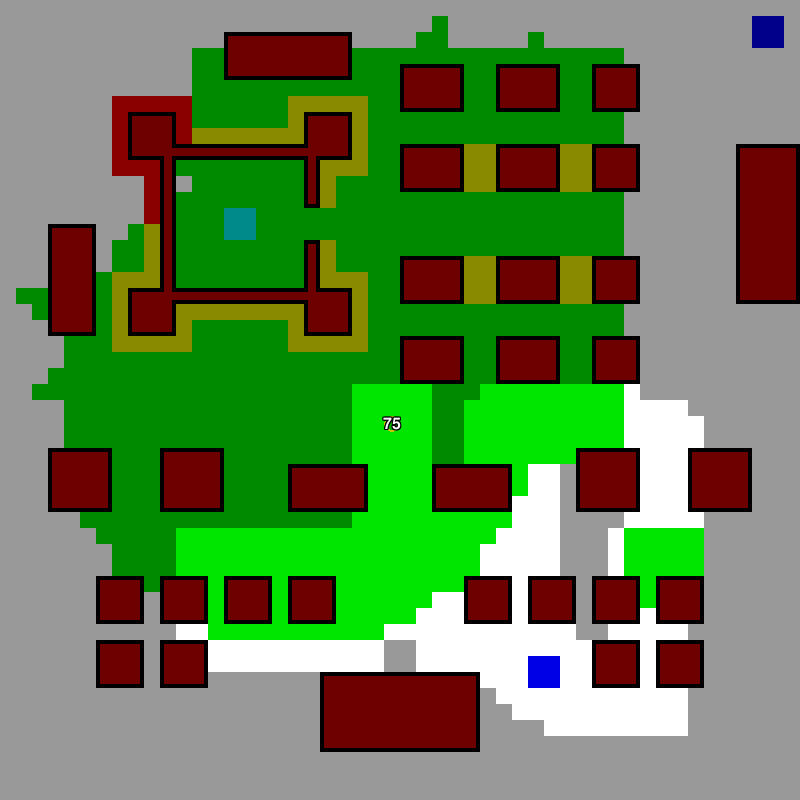}} & $\pi(\cdot | s)$ & \includegraphics[align=c,height=\pheight]{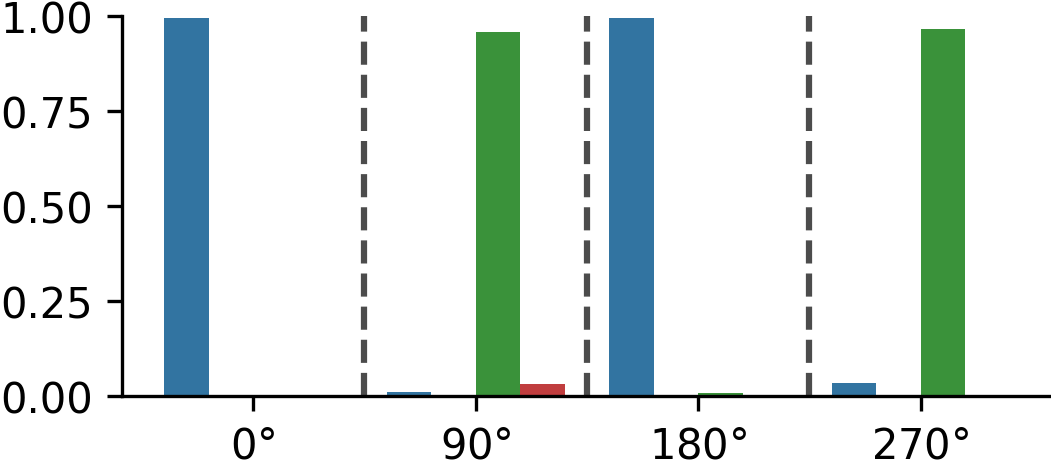} & \includegraphics[align=c,height=\pheight]{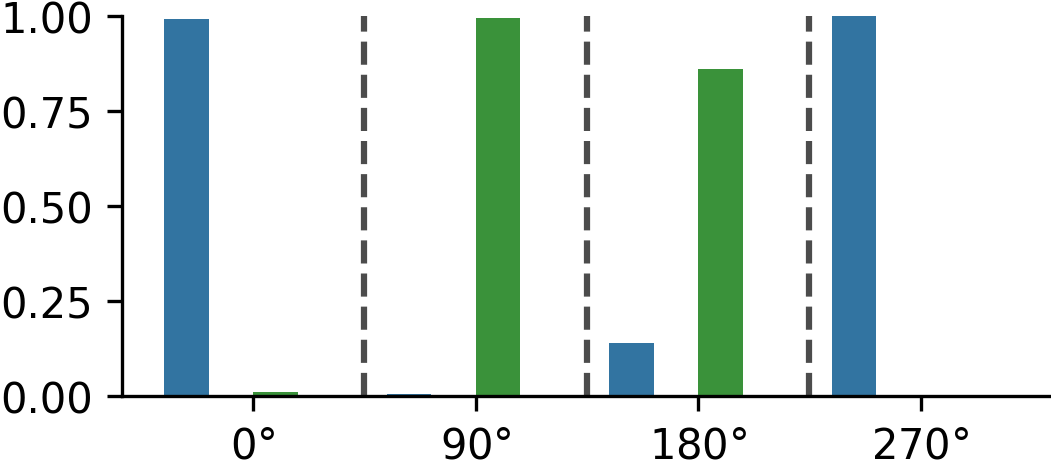}& \includegraphics[align=c,height=\pheight]{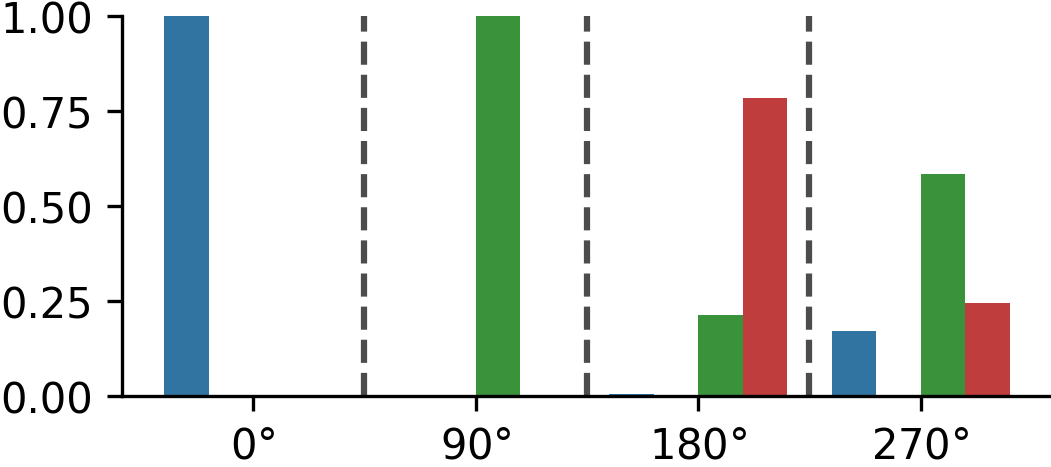}& \includegraphics[align=c,height=\pheight]{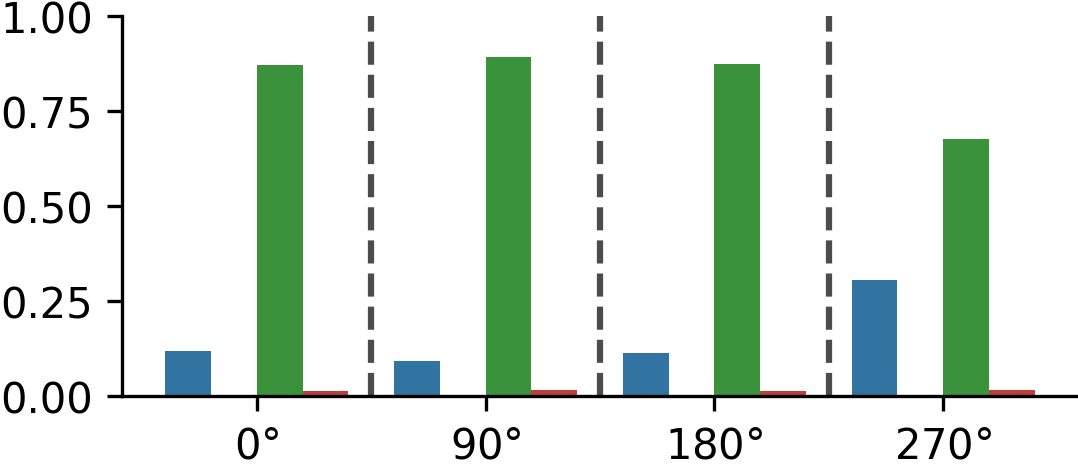}& \includegraphics[align=c,height=\pheight]{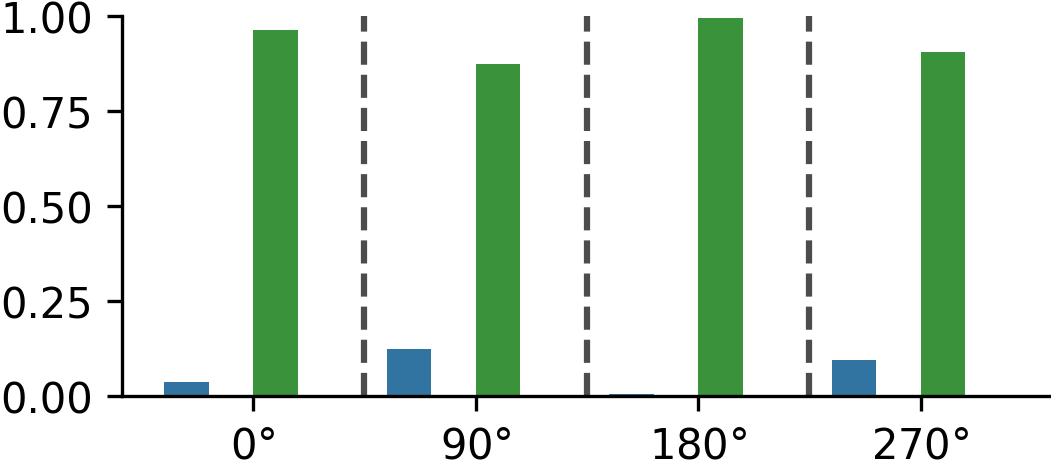}\bigstrut\\
       \cline{2-7}
         & $|\Delta\mathrm{V}(s)|$&[2.09 0.83 1.23 0.03 \!\!]&[0.25 0.07 0.17 0.15 \!\!]&[0.71 1.12 1.02 0.80 \!\!]&[0.01 0.02 0.03 0.02 \!\!]&[0.02 0.00 0.03 0.01 \!\!] \bigstrut\\
     \hline\hline
     \multirow{2}[2]{*}[3.5mm]{\includegraphics[width=1.5cm]{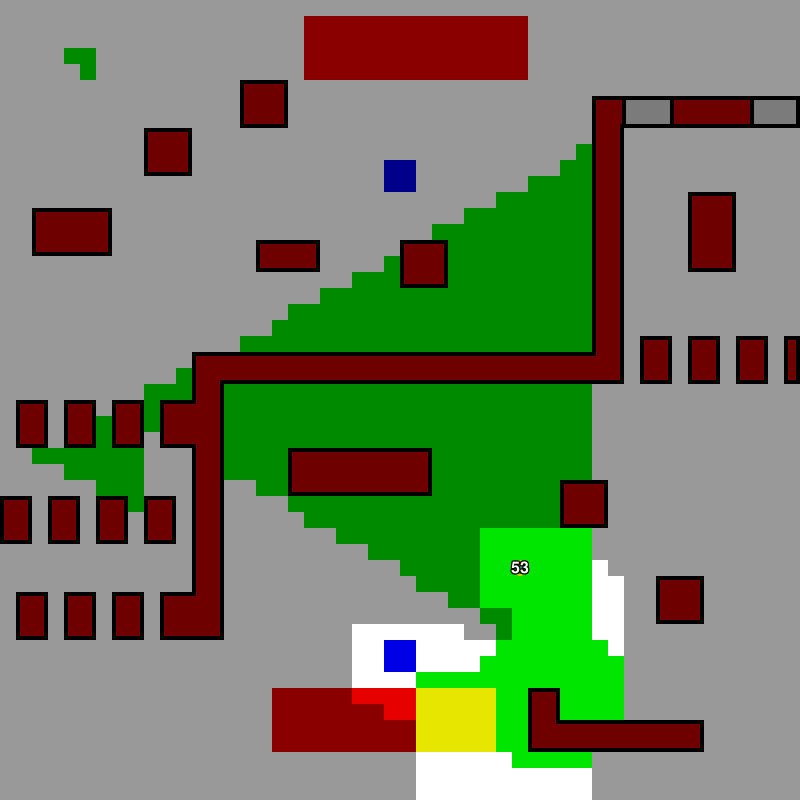}} & $\pi(\cdot | s)$ & \includegraphics[align=c,height=\pheight]{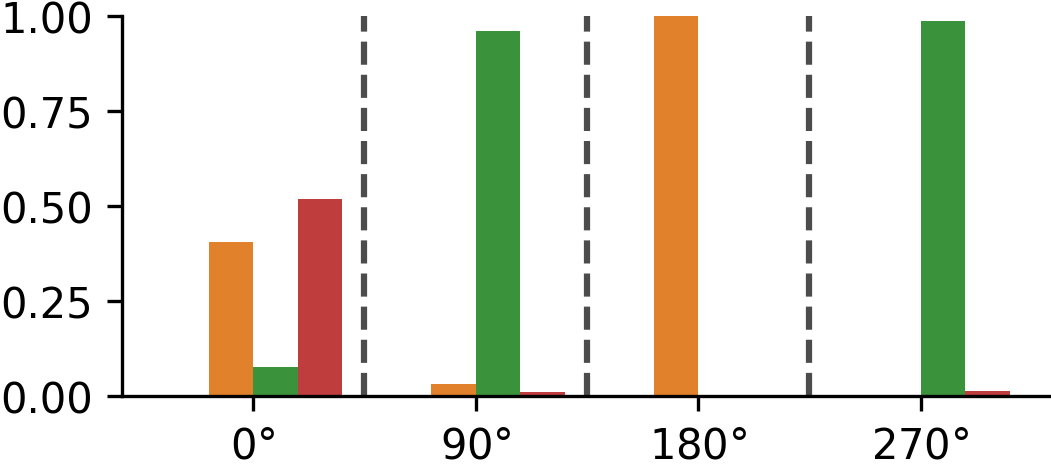} & \includegraphics[align=c,height=\pheight]{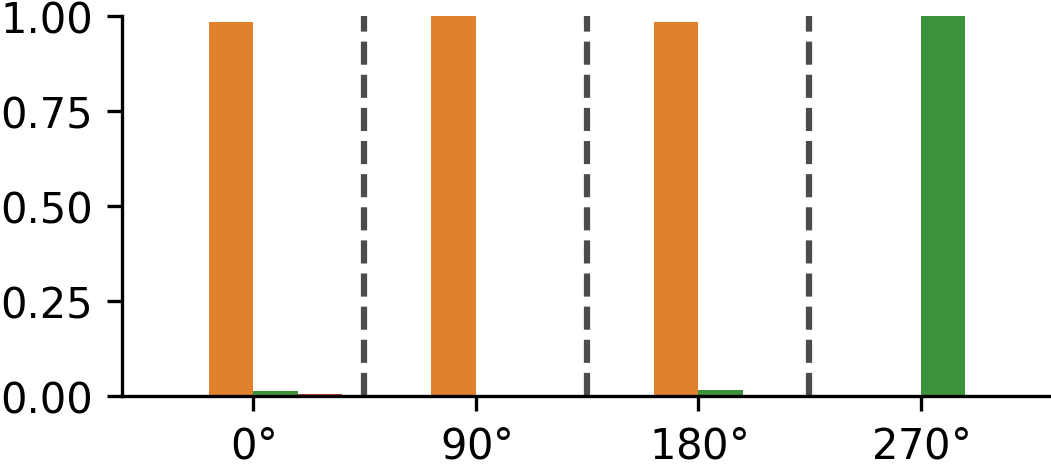}& \includegraphics[align=c,height=\pheight]{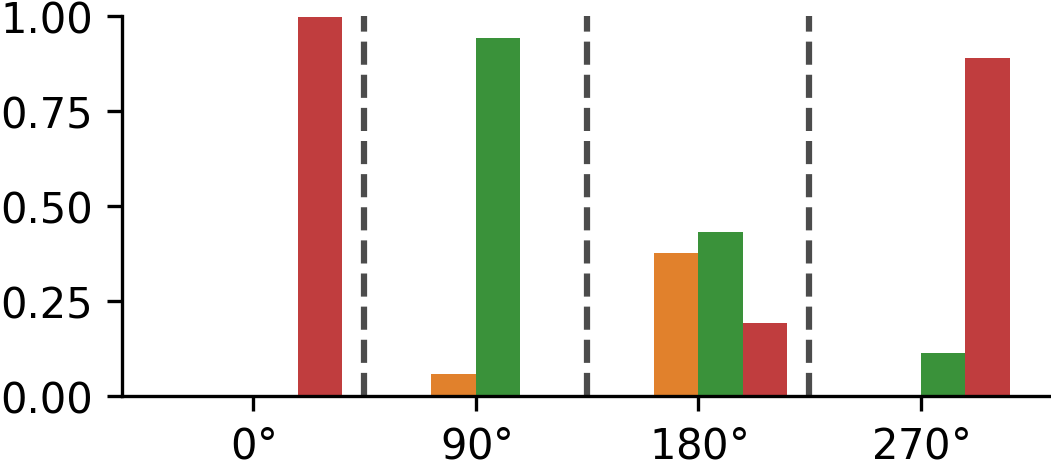}& \includegraphics[align=c,height=\pheight]{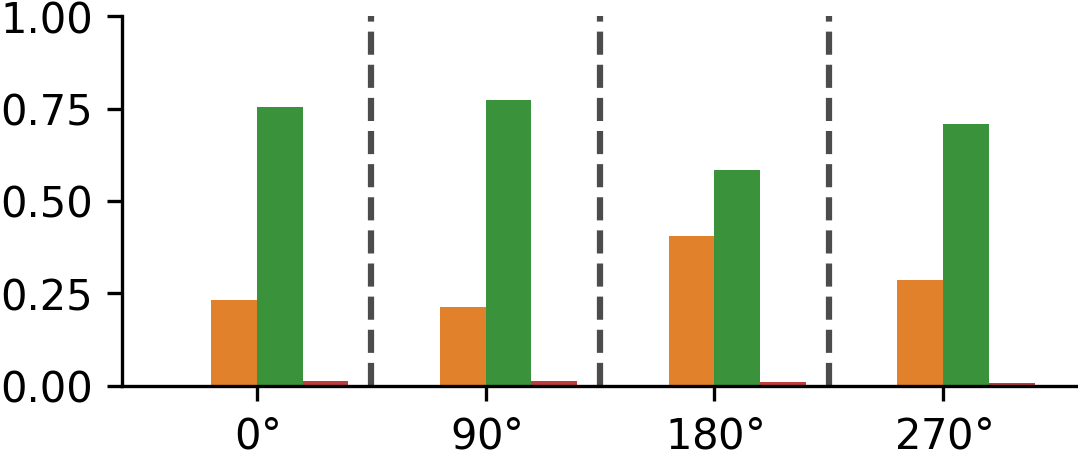}& \includegraphics[align=c,height=\pheight]{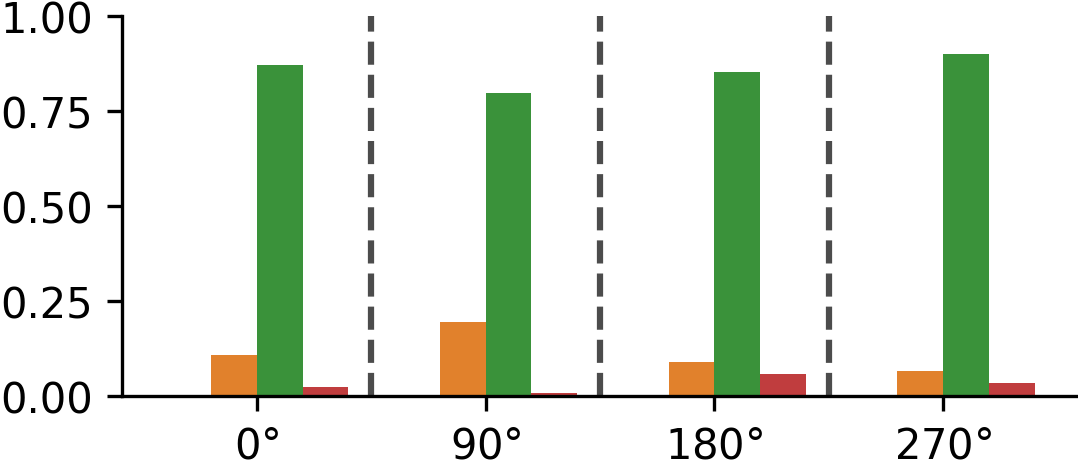}\bigstrut\\
       \cline{2-7}
         & $|\Delta\mathrm{V}(s)|$&[0.45 0.24 0.27 0.06 \!\!]&[0.24 0.19 0.44 0.49 \!\!]&[0.52 0.21 0.12 0.61 \!\!]&[0.15 0.16 0.10 0.11 \!\!]&[0.08 0.06 0.04 0.09 \!\!] \bigstrut\\
         \hline\hline
    \end{tabu}
    \caption{Qualitative visualization of equivariance and invariance of the different agents' policies and value functions for one in-distribution state (top) and one out-of-distribution state (bottom). The policy rows show the different action distributions over the \textit{directional} actions \textcolor{blue}{\textbf{east}}, \textcolor{orange}{\textbf{south}}, \textcolor{green}{\textbf{west}}, and \textcolor{red}{\textbf{north}} for all four rotations after transforming them back to the normal direction, such that an equivariant policy should show the same distribution for each rotation. The value row shows the absolute difference of the value estimate for each rotation to the mean, such that an invariant value function should show values at 0. }
    \label{tab:equiv}
\end{table*}

In Table~\ref{tab:comparison}, we show that the performance of the Regularized agent is lower for the rotated maps, indicating that it is not equivariant. To examine this in more detail, Table~\ref{tab:equiv} shows the action distribution and value estimation for all transformations and agents in two states, one in-distribution and one out-of-distribution. The action distribution is only shown for the directional actions, as the agent is not in a landing zone in both states, and thus the action mask is masking out the other three actions. In the table, an equivariant policy should show precisely the same distribution for the four rotations indicated through the 0\textdegree, 90\textdegree, 180\textdegree, and 270\textdegree. 

Focussing on the policies in both states, the Baseline, Augment, and Ensemble agents show significantly different action distributions. On the other hand, the Regularized and Ens.+Reg. agents show very similar distributions for each rotation, which can be seen for the in-distribution and out-of-distribution states. However, the distributions are not exactly the same, confirming that the added regularization loss does not create exact equivariant policies but rather ``nearly-equivariant'' ones. The same can be observed for the value estimate, which is shown as the absolute difference to the mean of the value estimate for each rotation. The agents with regularization produce values significantly closer to the mean than the others, showing that they are ``more invariant''.

This result yields the critical insight that regularization brings the policy toward being equivariant but not to exact equivariance. Since the value estimate should only be invariant if the policy is equivariant, it is essential to consider that the regularization of the critic, as done in \cite{raileanu2021automatic, nguyen2024symmetry}, may not be as effective as expected if the actor is only regularized and no ensemble is used.

\section{Discussion}

In this paper, we proposed the usage of equivariant ensembles and regularization to exploit symmetries in MDPs. We proved that ensemble policies and value functions are respectively equivariant and invariant, providing theoretical soundness to value regularization. We demonstrated the benefits of the proposed approach in a map-based path planning case study, showing a decrease in sample complexity and an increase in performance and generalization.

This paper assumes that symmetries in the MDP are perfect. However, this assumption may not be valid in some real-world applications. In asymmetrical cases, our approach may still work as long as the asymmetry is observable to the agent; this we will study in future work. The additional computational overhead needs to be considered, especially when the group of transformations grows large, e.g., by adding the flip symmetries in the case study. In that case, we would need to investigate whether we can randomly sample a subset of the transformations, which could yield equivariance in expectation. Furthermore, in this specific case study, we saw the limits on out-of-distribution generalization. These limits could be overcome by training the agent on procedurally generated maps, for which the equivariant ensembles and regularization would play an important role.

In future work, we will also investigate the effect of equivariant ensembles and regularization for different problem settings, specifically for continuous action spaces. Further, to make it more broadly applicable, we will study the effects when applying different RL algorithms such as the off-policy SAC algorithm~\cite{haarnoja2018soft}.

\bibliographystyle{IEEEtran}
\bibliography{IEEEabrv,bib}

% Generated by IEEEtran.bst, version: 1.14 (2015/08/26)
\begin{thebibliography}{10}
\providecommand{\url}[1]{#1}
\csname url@samestyle\endcsname
\providecommand{\newblock}{\relax}
\providecommand{\bibinfo}[2]{#2}
\providecommand{\BIBentrySTDinterwordspacing}{\spaceskip=0pt\relax}
\providecommand{\BIBentryALTinterwordstretchfactor}{4}
\providecommand{\BIBentryALTinterwordspacing}{\spaceskip=\fontdimen2\font plus
\BIBentryALTinterwordstretchfactor\fontdimen3\font minus \fontdimen4\font\relax}
\providecommand{\BIBforeignlanguage}[2]{{%
\expandafter\ifx\csname l@#1\endcsname\relax
\typeout{** WARNING: IEEEtran.bst: No hyphenation pattern has been}%
\typeout{** loaded for the language `#1'. Using the pattern for}%
\typeout{** the default language instead.}%
\else
\language=\csname l@#1\endcsname
\fi
#2}}
\providecommand{\BIBdecl}{\relax}
\BIBdecl

\bibitem{krizhevsky2012imagenet}
A.~Krizhevsky, I.~Sutskever, and G.~E. Hinton, ``Imagenet classification with deep convolutional neural networks,'' \emph{Advances in neural information processing systems}, vol.~25, 2012.

\bibitem{van2020mdp}
E.~Van~der Pol, D.~Worrall, H.~van Hoof, F.~Oliehoek, and M.~Welling, ``Mdp homomorphic networks: Group symmetries in reinforcement learning,'' \emph{Advances in Neural Information Processing Systems}, vol.~33, pp. 4199--4210, 2020.

\bibitem{wang2021mathrm}
D.~Wang, R.~Walters, and R.~Platt, ``So(2) equivariant reinforcement learning,'' in \emph{International Conference on Learning Representations}, 2021.

\bibitem{raileanu2021automatic}
R.~Raileanu, M.~Goldstein, D.~Yarats, I.~Kostrikov, and R.~Fergus, ``Automatic data augmentation for generalization in reinforcement learning,'' \emph{Advances in Neural Information Processing Systems}, vol.~34, pp. 5402--5415, 2021.

\bibitem{nguyen2024symmetry}
H.~Nguyen, T.~Kozuno, C.~C. Beltran-Hernandez, and M.~Hamaya, ``Symmetry-aware reinforcement learning for robotic assembly under partial observability with a soft wrist,'' \emph{arXiv preprint arXiv:2402.18002}, 2024.

\bibitem{schulman2017proximal}
J.~Schulman, F.~Wolski, P.~Dhariwal, A.~Radford, and O.~Klimov, ``Proximal policy optimization algorithms,'' \emph{arXiv preprint arXiv:1707.06347}, 2017.

\bibitem{theile2023learning}
M.~Theile, H.~Bayerlein, M.~Caccamo, and A.~L. Sangiovanni-Vincentelli, ``Learning to recharge: {UAV} coverage path planning through deep reinforcement learning,'' \emph{arXiv preprint arXiv:2309.03157}, 2023.

\bibitem{theile2020uav}
M.~Theile, H.~Bayerlein, R.~Nai, D.~Gesbert, and M.~Caccamo, ``Uav coverage path planning under varying power constraints using deep reinforcement learning,'' in \emph{2020 IEEE/RSJ International Conference on Intelligent Robots and Systems (IROS)}.\hskip 1em plus 0.5em minus 0.4em\relax IEEE, 2020, pp. 1444--1449.

\bibitem{sutton1998introduction}
R.~S. Sutton and A.~G. Barto, \emph{Reinforcement Learning: {A}n Introduction}.\hskip 1em plus 0.5em minus 0.4em\relax Cambridge, MA: The MIT Press, 1998.

\bibitem{schulman2015high}
J.~Schulman, P.~Moritz, S.~Levine, M.~Jordan, and P.~Abbeel, ``High-dimensional continuous control using generalized advantage estimation,'' \emph{arXiv preprint arXiv:1506.02438}, 2015.

\bibitem{cohen2019general}
T.~S. Cohen, M.~Geiger, and M.~Weiler, ``A general theory of equivariant cnns on homogeneous spaces,'' \emph{Advances in neural information processing systems}, vol.~32, 2019.

\bibitem{finzi2021residual}
M.~Finzi, G.~Benton, and A.~G. Wilson, ``Residual pathway priors for soft equivariance constraints,'' \emph{Advances in Neural Information Processing Systems}, vol.~34, pp. 30\,037--30\,049, 2021.

\bibitem{wang2022equivariant}
D.~Wang, R.~Walters, X.~Zhu, and R.~Platt, ``Equivariant $ q $ learning in spatial action spaces,'' in \emph{Conference on Robot Learning}.\hskip 1em plus 0.5em minus 0.4em\relax PMLR, 2022, pp. 1713--1723.

\bibitem{wang2022surprising}
D.~Wang, J.~Y. Park, N.~Sortur, L.~L. Wong, R.~Walters, and R.~Platt, ``The surprising effectiveness of equivariant models in domains with latent symmetry,'' in \emph{The Eleventh International Conference on Learning Representations}, 2022.

\bibitem{nguyen2023equivariant}
H.~H. Nguyen, A.~Baisero, D.~Klee, D.~Wang, R.~Platt, and C.~Amato, ``Equivariant reinforcement learning under partial observability,'' in \emph{Conference on Robot Learning}.\hskip 1em plus 0.5em minus 0.4em\relax PMLR, 2023, pp. 3309--3320.

\bibitem{yarats2020image}
D.~Yarats, I.~Kostrikov, and R.~Fergus, ``Image augmentation is all you need: Regularizing deep reinforcement learning from pixels,'' in \emph{International conference on learning representations}, 2020.

\bibitem{laskin2020reinforcement}
M.~Laskin, K.~Lee, A.~Stooke, L.~Pinto, P.~Abbeel, and A.~Srinivas, ``Reinforcement learning with augmented data,'' \emph{Advances in neural information processing systems}, vol.~33, pp. 19\,884--19\,895, 2020.

\bibitem{haarnoja2018soft}
T.~Haarnoja, A.~Zhou, P.~Abbeel, and S.~Levine, ``Soft actor-critic: Off-policy maximum entropy deep reinforcement learning with a stochastic actor,'' in \emph{International conference on machine learning}.\hskip 1em plus 0.5em minus 0.4em\relax PMLR, 2018, pp. 1861--1870.

\bibitem{arkin2000approximation}
E.~M. Arkin, S.~P. Fekete, and J.~S. Mitchell, ``Approximation algorithms for lawn mowing and milling,'' \emph{Computational Geometry}, vol.~17, no. 1-2, pp. 25--50, 2000.

\end{thebibliography}

\balance

\end{document}